%% file: main_arxiv.tex
\documentclass[11pt]{article}

\usepackage{microtype}
\usepackage{graphicx}
\usepackage{booktabs} % for professional tables

\usepackage{natbib}
\usepackage{fullpage}
\usepackage[utf8]{inputenc}
\usepackage[T1]{fontenc}
\usepackage{microtype}
\usepackage{csquotes}
\usepackage[osf,sc]{mathpazo}
\RequirePackage[scaled=0.90]{helvet}
\RequirePackage[scaled=0.85]{beramono}
\RequirePackage{textcomp}
\usepackage{xcolor}
\definecolor{DarkGreen}{rgb}{0.075,0.375,0.075}
\definecolor{DarkRed}{rgb}{0.5,0.1,0.1}
\definecolor{DarkBlue}{rgb}{0.1,0.1,0.5}
\definecolor{Gray}{rgb}{0.2,0.2,0.2}
\usepackage[pdftex]{hyperref}
\hypersetup{
    unicode=false,          % non-Latin characters in AcrobatÂ¿s bookmarks
    pdftoolbar=true,        % show Acrobat toolbar?
    pdfmenubar=true,        % show Acrobat menu?
    pdffitwindow=false,      % page fit to window when opened
    pdfnewwindow=true,      % links in new window
    colorlinks=true,       % false: boxed links; true: colored links
    linkcolor=DarkBlue,          % color of internal links
    citecolor=DarkGreen,        % color of links to bibliography
    filecolor=DarkRed,      % color of file links
    urlcolor=DarkBlue,          % color of external links
    pdftitle={},
    pdfauthor={},
}
\usepackage[skip=2mm,indent=5mm]{parskip}
\usepackage[small]{caption}
\RequirePackage{amsthm,amsmath,amsfonts,amssymb}

\usepackage[capitalize,noabbrev]{cleveref}

\usepackage[utf8]{inputenc} % allow utf-8 input
\usepackage[T1]{fontenc}    % use 8-bit T1 fonts
\usepackage{hyperref}       % hyperlinks
\usepackage{url}            % simple URL typesetting
\usepackage{booktabs}       % professional-quality tables
\usepackage{amsfonts}       % blackboard math symbols
\usepackage{nicefrac}       % compact symbols for 1/2, etc.
\usepackage{microtype}      % microtypography
\usepackage{xcolor}         % colors
\usepackage{amsmath}
\usepackage{amsthm} %Theorems
\usepackage{enumitem}
\usepackage{pifont}
\usepackage{wrapfig}
\usepackage{bm}
\usepackage{bbm}
\usepackage{todonotes}
\usepackage{makecell}

\usepackage{caption}
\usepackage{subcaption}

\input{math_commands}

\usepackage[ruled,vlined]{algorithm2e}
\usepackage{etoolbox}

\newbool{show_abstract_dagger}
\setbool{show_abstract_dagger}{true}
\newbool{show_model_table}
\setbool{show_model_table}{true}
\newbool{show_full_size_fig}
\setbool{show_full_size_fig}{true}
\newbool{use_citet}
\setbool{use_citet}{false}
\newbool{transpose_banner}
\setbool{transpose_banner}{false}

\title{Train-before-Test Harmonizes Language Model Rankings}

\author{%
  Guanhua Zhang\footnote{Corresponding author: \href{mailto:guanhua.zhang@tuebingen.mpg.de}{guanhua.zhang@tuebingen.mpg.de}}, Ricardo Dominguez-Olmedo, Moritz Hardt
}
\date{
\textit{Max Planck Institute for Intelligent Systems, Tübingen} and \textit{Tübingen AI Center}\\
}

\newcounter{daggerfootnote}
\newcommand*{\daggerfootnote}[1]{%
    \setcounter{daggerfootnote}{\value{footnote}}%
    \renewcommand*{\thefootnote}{\fnsymbol{footnote}}%
    \footnote[2]{#1}%
    \setcounter{footnote}{\value{daggerfootnote}}%
    \renewcommand*{\thefootnote}{\arabic{footnote}}%
    }

\begin{document}

\onecolumn
\maketitle

\input{sections/abstract}
\input{sections/intro}

\input{sections/related}
\input{sections/experiments}

\input{sections/discussion}

\input{sections/ackownledge}

% \newpage
\bibliographystyle{plain}
\bibliography{ref}
\newpage
\input{sections/appendix}

\end{document}

%% file: math_commands.tex
\def\eqref#1{equation~\ref{#1}}
\def\1{\bm{1}}

\DeclareMathAlphabet{\mathsfit}{\encodingdefault}{\sfdefault}{m}{sl}
\SetMathAlphabet{\mathsfit}{bold}{\encodingdefault}{\sfdefault}{bx}{n}

%% file: sections/abstract.tex
\begin{abstract}
Existing language model benchmarks provide contradictory model rankings, even for benchmarks that aim to capture similar skills. 
This dilemma of conflicting rankings hampers model selection, clouds model comparisons, and adds confusion to a growing ecosystem of competing models.
In this paper, we take a different perspective on model comparison: instead of relying on out-of-the-box performance via direct evaluation, we compare \emph{model potential} by providing each model with identical benchmark-specific fine-tuning before evaluation.
We call this approach \emph{train-before-test}.
Our primary contribution is a comprehensive empirical evaluation of model potential across 24 benchmarks and 61 models.
First, we demonstrate that model potential rankings obtained through train-before-test exhibit remarkable consistency across all benchmarks. 
Whereas traditional rankings demonstrate little external validity under direct evaluation, they enjoy a significant degree of external validity when applying train-before-test: model potential rankings transfer gracefully from one benchmark to another. 
Second, train-before-test restores the connection between perplexity and downstream task performance, lost under direct evaluation.
Remarkably, even pre-finetuning perplexity of a base model predicts post-finetuning downstream performance, suggesting that ranking consistency reflects inherent model potential rather than fine-tuning artifacts.
Finally, train-before-test reduces the model-score matrix to essentially rank one, indicating that model potential is dominated by one latent factor, uncovered by train-before-test.
Our work supports the recommendation to make train-before-test a default component of LLM benchmarking\ifbool{show_abstract_dagger}{\daggerfootnote{Code is available at \url{https://github.com/socialfoundations/lm-harmony}.}}{}.

% Existing language model benchmarks provide contradictory model rankings, even for benchmarks that aim to capture similar skills. This dilemma of conflicting rankings hampers model selection, clouds model comparisons, and adds confusion to a growing ecosystem of competing models. Recent work attributed ranking disagreement to the phenomenon of \emph{training on the test task}: As released, different models exhibit a different level of preparation for any given test task. A candidate solution to the problem is \emph{train-before-test}: Give each model the same benchmark-specific finetuning before evaluation. Our primary contribution is a broad empirical evaluation of train-before-test across 24 benchmarks and 61 models. We show that train-before-test significantly improves ranking agreement consistently across all benchmarks. Whereas rankings have little external validity to start with, they enjoy a significant degree of external validity when applying train-before-test: Model rankings transfer gracefully from one benchmark to the other. Even within the same model family, train-before-test reduces strong ranking disagreement to near-perfect agreement. In addition, train-before-test reduces the model-score matrix to essentially rank one, revealing new insights into the latent factors of benchmark performance. Our work supports the recommendation to make train-before-test a default component of LLM benchmarking\daggerfootnote{Code is available at \url{https://github.com/socialfoundations/lm-harmony}.}.

\end{abstract}

%% file: sections/intro.tex
\section{Introduction}
\label{sec:intro}

Existing language model benchmarks provide contradictory model rankings, even for benchmarks that aim to capture similar skills~\citep{liang2022holistic,open-llm-leaderboard,open-llm-leaderboard-v2}. 
This inconsistency poses a serious challenge: how can we reliably compare, rank, and select models when different benchmarks yield conflicting information? While this ranking disagreement is often attributed to the diverse capabilities of large language models~\citep{Ruan2024ObservationalSL}, it creates a conundrum in practice that muddles model development decisions~\citep{zhang2024inherent}.

Current evaluation methodology works from \emph{direct evaluation}, probing models via black-box function calls. However, large language models are trained on diverse, often proprietary data mixes that vary significantly across models~\citep{llama,gemma,Guha2025OpenThoughtsDR}. 
Recent work showed that this leads to the problem of \emph{training on the test task~\citep{dominguez2024training}:} the extent to which a model has encountered data similar to the test task during training confounds model comparisons, rankings, and scaling laws \cite{kaplan2020scaling}. Put simply, an otherwise inferior model may have simply prepared better for a specific task. 

In this paper, we take a fresh perspective on evaluation methodology: in contrast with direct evaluation, we compare \emph{model potential} by giving each model the same task-specific fine-tuning. 
We call this approach \emph{train-before-test}. Its goal is to achieve valid model comparisons by ensuring that all models receive equal preparation for the test.

We envision train-before-test as a tool for \emph{regret-free} model selection for downstream applications. Increasingly, practitioners select one from many available models with the goal of adapting for a specific task. Under direct evaluation the best model to begin with may no longer be the best model after task-specific preparation. In contrast, we show that train-before-task yields model comparisons and rankings that enjoy broad external validity.
% In more detail, we make the following key contributions.

\subsection{Our Contributions}
\ifbool{transpose_banner}{
\begin{figure}
    \centering
    \includegraphics[width=0.99\textwidth]{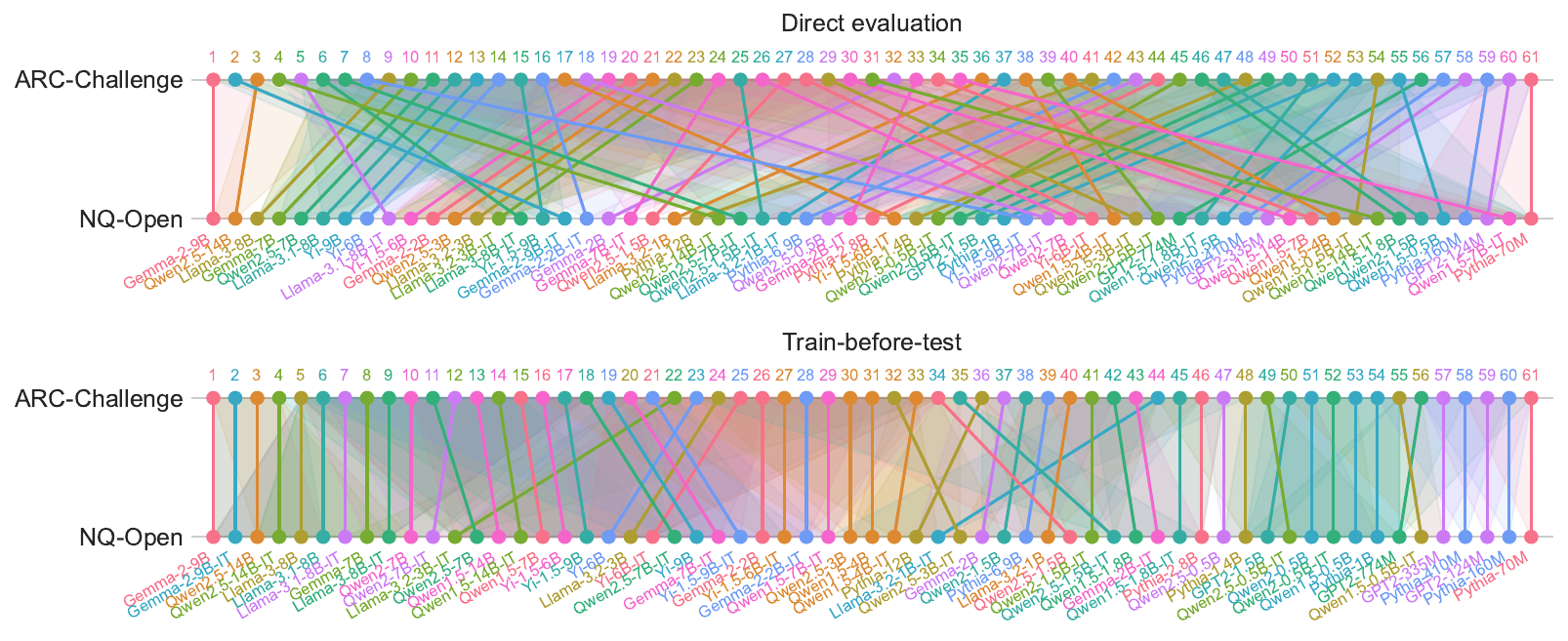}
    \caption{Rankings of 61 language models on two question-answering benchmarks: Natural Questions Open and ARC Challenge. {\bf Top:} Direct evaluation leads to inconsistent rankings. Although both benchmarks test for question-answering ability, the resulting model rankings show substantial disagreement. {\bf Bottom:} Train-before-test aligns model rankings. 
    {\bf Note:} For each of the two plots, we greedily align model rankings as much as possible without violating confidence intervals, thus revealing only those ranking changes that are statistically significant.
    See Appendix~\ref{app:banner}.
    }
    \label{fig:banner}
\end{figure}
}{
\begin{figure}
    \centering
    \includegraphics[width=0.99\textwidth]{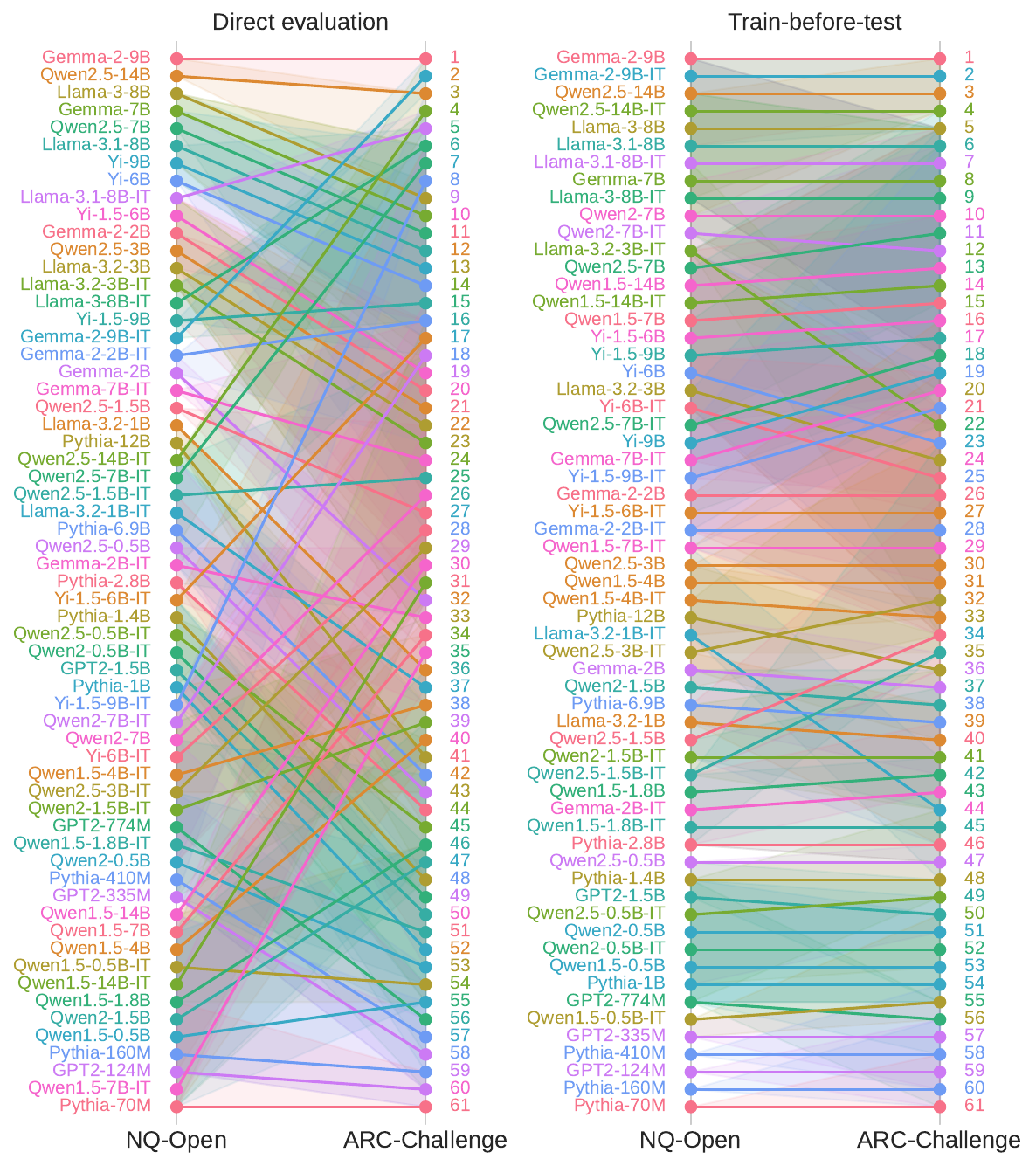}
    \caption{Rankings of 61 language models on two question-answering benchmarks: Natural Questions Open and ARC Challenge. {\bf Left:} Direct evaluation leads to inconsistent rankings. Although both benchmarks test for question-answering ability, the resulting model rankings show substantial disagreement. {\bf Right:} Train-before-test aligns model rankings. 
    {\bf Note:} For each of the two plots, we greedily align model rankings as much as possible without violating confidence intervals, thus revealing only those ranking changes that are statistically significant.
    See Appendix~\ref{app:banner} for more details.
    }
    \label{fig:banner}
\end{figure}
}

\paragraph{Direct evaluation leads to ranking disagreement even between related tasks.}
We demonstrate that the prevalent direct evaluation scheme results in strong disagreement between model ranking across various benchmarks. We show that this strong ranking disagreement persists even when restricting to benchmarks that aim to capture similar tasks. Moreover, rankings still strongly disagree when evaluating models from the same family.
The situation presents a serious conundrum for model selection: Under direct evaluation, benchmarks fail to give reliable and actionable insights for model choosing among multiple alternatives.

\paragraph{Train-before-test leads to consistent model potential rankings.}
We comprehensively evaluate train-before-test across 24 benchmark datasets and 61 large language models. 
By fine-tuning each model on identical task-relevant data before evaluation, we uncover remarkably consistent model potential rankings. 
Ranking agreement between benchmarks, measured by Kendall's tau, improves for 274 out of 276 benchmark pairs, with the average Kendall's $\tau$ increasing from 0.52 to 0.76. 
Figure~\ref{fig:banner} illustrates the result for one typical pair of benchmarks.
This consistency suggests that model potential, unlike out-of-the-box performance, has external validity~\citep{Salaudeen2025MeasurementTM} and transfers gracefully across different tasks.

\paragraph{Model potential aligns perplexity rankings with downstream tasks.}

Perplexity benchmarks used to be popular, but fell out of fashion because of the apparent disconnect between perplexity and downstream task performance~\citep{weiemergent, ganguli2022predictability, liu2023same,Magnusson2023PalomaAB,Lourie2025ScalingLA}. 
We indeed validate this disconnect when comparing model families under direct evaluation. 
However, train-before-test restores this fundamental relationship in two ways.
First, we show that post-fine-tuning perplexity rankings align well with post-fine-tuning downstream task rankings, creating consistency between training objectives and task performance.
Second, and more remarkably, for base (non-instruction-tuned) models, even pre-fine-tuning perplexity predicts post-fine-tuning downstream performance.
This suggests that the ranking consistency we observe reflects inherent model potential rather than artifacts of fine-tuning.

\paragraph{Train-before-test sheds light on the latent factors of benchmark scores.}
Consider the large benchmark-model score matrix, where each entry $(i, j)$ corresponds to the performance of model~$j$ on a benchmark~$i$. Several works have considered this matrix for different reasons and found that it is approximately low rank~\citep{Ruan2024ObservationalSL,Owen2024HowPI,Burnell2023RevealingTS}, but not quite. The first singular value is dominant and correlates with pre-training compute. However, the other components aren’t negligible, and their interpretation remains unclear.
We show that train-before-test clarifies this state of affairs. After train-before-test, the benchmark-model matrix is essentially rank one. The first principal component accounts for $86\%$ of the explained variance across all models, and for $93\%$ of the variance for a single model family. 
This suggests that model potential is dominated by a single latent factor, while the additional components observed in direct evaluation may reflect task-specific training exposure.

%% file: sections/related.tex
\section{Related Work}
Benchmarking has played a central role in the advancement of machine learning~\citep{liberman2010obituary, hardt2022patterns}.
While absolute model performance is often fragile to even seemingly minor changes in evaluation data~\citep{candela2009dataset, torralba2011unbiased, albadawy2018deep, taori2020measuring, Tsipras2020FromIT}, relative model performance—that is, model rankings—tends to transfer surprisingly well across classical benchmarks~\citep{Yadav2019ColdCT,recht2019imagenetclassifiersgeneralizeimagenet, miller2020effect}. 
For instance, prior work~\citep{kornblith2019better, barbu2019objectnet} has shown that model rankings on ImageNet~\citep{deng2009imagenet} also transfer to other image classification and object recognition benchmarks. 
Moreover, \ifbool{use_citet}{\citet{salaudeen2024imagenot}}{Salaudeen and Hardt~(2024, \citep{salaudeen2024imagenot})} demonstrated that ImageNet rankings remain robust even under major dataset variations. 
This transferability of model rankings is highly desirable, as it indicates that progress on specific benchmarks reliably reflects broader scientific advancements~\citep{liao2021we, hardt2025emerging}.

However, the emergence of foundation models has dramatically transformed the benchmarking landscape compared to the ImageNet era~\citep{liang2022holistic, srivastava2022beyond, Weidinger2025TowardAE}.
With huge training costs and much improved capabilities~\citep{qwen,llama,Ramesh2021ZeroShotTG,Anil2023GeminiAF,openai2023GPT4}, practitioners now lean towards directly evaluating LLMs across a wide range of different benchmarks, in the hope of obtaining a more comprehensive assessment of their capabilities~\citep{liang2022holistic,bbh,mmlu,open-llm-leaderboard,open-llm-leaderboard-v2}.
This shift introduces new challenges, as model rankings across different tasks may vary significantly~\citep{huan2025does,lourie2025scaling}.
\ifbool{use_citet}{\citet{zhang2024inherent}}{Zhang and Hardt~(2024, \citep{zhang2024inherent})} draw an analogy between multi-task benchmarks and voting systems~\citep{arrow2012social}, revealing that a multi-task benchmarking approach with diverse rankings inherently lacks robustness to minor changes and thus cannot provide a stable unified ranking. 

This lack of unified ranking is sometimes seen as a desirable feature within the community~\citep{liang2022holistic}.
Some argue that variability reflects the multifaceted strengths and weaknesses of LLMs, suggesting that users should select the best model tailored to their specific needs~\citep{ghosh2024onebench,Zhang2024TaskMA,Shnitzer2023LargeLM}.
For example, a user who focuses on mathematical tasks could prioritize the math benchmark to choose the optimum model. 
However, there are two significant concerns regarding this approach:
First, the user-driven selection strategy poses challenges for model developers. 
Given the resource-intensive nature of LLM development~\citep{guo2025deepseek}, it is impractical to release a different model for every potential use case.
Moreover, developers typically aim to create a general-purpose model~\citep{qwen,llama}; however, such a desideratum is often difficult to reliably measure due to the inconsistent rankings observed across benchmarks.
Second, we demonstrate in this paper that benchmarks within the same task category can still exhibit substantial discrepancies in model rankings. 

One potential reason for the observed inconsistencies in model rankings is that models vary substantially in their training data~\citep{Gadre2023DataCompIS, albalaksurvey}.
In particular, \ifbool{use_citet}{\citet{dominguez2024training}}{Dominguez et al.~(2024, \citep{dominguez2024training})} show that models vary in their degree of preparedness for popular benchmarks. 
Building on this idea, we introduce the notion of train-before-test, wherein we fine-tune each model on the corresponding training set so every model arrives well prepared.
While extensive literature exists on investigating different fine-tuning strategies for LLMs~\citep{Lin2024SelectingLL,Zeng2025LENSLLMUF,Lester2021ThePO}, this lies outside the scope of our investigation. 
Instead, we apply standardized fine-tuning~\citep{peft} as an evaluation tool to give all models equivalent preparation before testing.
Rather than studying models from the same family with varying pre-training compute~\citep{Zhang2024WhenSM,kaplan2020scaling}, our experiments cover 61 models from six families and 24 tasks from different categories.
We study how train-before-test improves ranking consistency across benchmarks and its implications for benchmarking practices.

%% file: sections/experiments.tex
\section{Experiments}
\label{sec:exp}
In this section, we examine the cross-benchmark ranking agreement of 61 language models across 24 benchmarks. We find that ranking agreement tends to be low, with an average Kendall's $\tau$ of 0.52. We then examine benchmark agreement under a different benchmarking methodology, which we refer to as \emph{train-before-test}. Specifically, we fine-tune on a benchmark's train set prior to evaluating on said benchmark. Compared to direct evaluation, train-before-test improves cross-benchmark ranking agreement on almost all benchmark pairs considered. The improvements in ranking agreement are typically large, with the average Kendall's $\tau$ increasing to 0.76. 

We additionally find that train-before-test greatly improves the agreement between perplexity rankings and downstream benchmarks. This result holds consistently across three general domain corpus, newly collected from \texttt{Wiki}, \texttt{arXiv}, and \texttt{Stack}Exchange.
We retrained content only from 2025 to ensure models had not seen those texts during pretraining.
The average Kendall's $\tau$ between perplexity ranking and 24 downstream task rankings improves from 0.48 to 0.74 with train-before-test, leading to much better consistency between the training objective and downstream benchmark performance.
Moreover, for base models, pre-fine-tuning perplexity ranking remains consistent with post-fine-tuning downstream rankings (average Kendall’s $\tau$ = 0.78). 
This consistency does not hold for instruction-tuned models.

Finally, we discuss the implications of increased cross-benchmark agreement. In doing so, we perform Principal Component Analysis (PCA) over the model score matrix comprising all benchmark scores. We then analyze its principal components both under direct model evaluation and train-before-test. We find that train-before-test greatly increases the share of variance explained by the first principal component (PC1), from 70\% for direct evaluation to 86\% for train-before-test. In both cases, PC1 aligns well with model pre-training compute. By further controlling for model family and only considering \textsc{Qwen} models, we show that the explained variance ratio of PC1 further increases to 93\%, making the model-score matrix essentially rank one.

\subsection{Experiment Setting}
\label{sec:setting}

\ifbool{use_citet}{
\begin{wraptable}{r}{0.55\textwidth}
\vspace{-4.5mm}
% \begin{table}[t]
\caption{ 
We categorize benchmarks into language understanding (LU), commonsense reasoning (CR), question answering (QA), physics/biology/chemistry (PBC), math (Math), and medicine (Med).
}
\label{tab:benchmark_categories}
\centering
\resizebox{\linewidth}{!}{
\begin{tabular}{l|l}
\toprule
{Category}                         & {Benchmarks} \\ 
\midrule
LU       & \begin{tabular}[c]{@{}l@{}}\texttt{MNLI}, \texttt{QNLI}, \texttt{RTE}, \texttt{CoLA}, \texttt{SST-2}, \texttt{MRPC}, \texttt{QQP}, \texttt{WiC}, \texttt{ANLI}\end{tabular} \\[0.4em] %\hline
CR                      & \begin{tabular}[c]{@{}l@{}}\texttt{Winogrande}, \texttt{CommonsenseQA},  \texttt{Hellaswag}, \texttt{Social-IQA} \end{tabular} \\[0.4em] %\hline
QA                        & \begin{tabular}[c]{@{}l@{}} \texttt{OpenBookQA}, \texttt{NQ-Open}, \texttt{BoolQ}, \texttt{ARC-Easy}, \texttt{ARC-Challenge} \end{tabular} \\[0.4em] %\hline
 PBC    & \texttt{SciQ}, \texttt{PIQA}  \\[0.4em] %\hline
Math & \texttt{MathQA}, \texttt{GSM8K} \\[0.4em] %\hline
Med & \texttt{MedMCQA}, \texttt{HeadQA} \\[0.4em]
% \hline
% \midrule
\bottomrule
\end{tabular}}
% \end{table}
\vspace{-5mm}
\end{wraptable}
}{
\begin{table}[t]
\caption{ 
We categorize benchmarks into language understanding (LU), commonsense reasoning (CR), question answering (QA), physics/biology/chemistry (PBC), math (Math), and medicine (Med).
}
\label{tab:benchmark_categories}
\centering
\resizebox{1.0\textwidth}{!}{
\begin{tabular}{l|l}
\toprule
{Category}                         & {Benchmarks} \\ 
\midrule
LU       & \begin{tabular}[c]{@{}l@{}}\texttt{MNLI}~\citep{mnli}, \texttt{QNLI}~\citep{qnli}, \texttt{RTE}~\citep{rte,rte2,rte3}, \texttt{CoLA}~\citep{cola}, \texttt{SST-2}~\citep{sst2}, \texttt{MRPC}~\citep{mrpc}, \texttt{QQP}, \texttt{WiC}~\citep{wic}, \texttt{ANLI}~\citep{anli}\end{tabular} \\[0.4em] %\hline
CR                      & \begin{tabular}[c]{@{}l@{}}\texttt{Winogrande}~\citep{winograd}, \texttt{CommonsenseQA}~\citep{commonsenseqa},  \texttt{Hellaswag}~\citep{hellaswag}, \texttt{Social-IQA}~\citep{socialiqa} \end{tabular} \\[0.4em] %\hline
QA                        & \begin{tabular}[c]{@{}l@{}} \texttt{OpenBookQA}~\citep{openbookqa}, \texttt{NQ-Open}~\citep{nq_open}, \texttt{BoolQ}~\citep{boolq}, \texttt{ARC-Easy}, \texttt{ARC-Challenge}~\citep{allenai:arc} \end{tabular} \\[0.4em] %\hline
 PBC    & \texttt{SciQ}~\citep{sciq}, \texttt{PIQA}~\citep{piqa}  \\[0.4em] %\hline
Math & \texttt{MathQA}~\citep{mathqa}, \texttt{GSM8K}~\citep{gsm8k} \\[0.4em] %\hline
Med & \texttt{MedMCQA}~\citep{medmcqa}, \texttt{HeadQA}~\citep{headqa} \\[0.4em]
% \hline
% \midrule
\bottomrule
\end{tabular}}
\end{table}
}
\paragraph{Benchmark selection.} 
We begin our study with the \texttt{lm-eval-harness} package~\citep{eval-harness}, which offers a comprehensive suite of language model benchmarks. 
To accommodate the train-before-test methodology which requires a dedicated training set for fine-tuning, we first identify benchmarks that provide at least 1,000 training examples. 
This yields a total of 37 benchmarks, which we broadly categorize into 28 likelihood-based and 9 generation-based benchmarks.

Generation-based benchmarks are often computationally intensive to evaluate, as base models typically generate text until reaching their maximum sequence length. 
These benchmarks are also over-challenging for smaller models with limited parameters, such as \textsc{GPT-2}~\citep{gpt2}.
Therefore, we select only \texttt{NQ-Open} and \texttt{GSM8K} from the generation-based benchmarks.
Among the likelihood-based benchmarks, we further exclude six due to observed anomalies during fine-tuning, such as a lack of performance improvement in over 20\% of models.
See Appendix~\ref{app:data_setting} for details. 

Our final selection consists of 24 benchmarks covering diverse domains and task types. 
These benchmarks are primarily multiple-choice question answering benchmarks, with accuracy as the task metric. 
We categorize all benchmarks by their descriptions, see Table~\ref{tab:benchmark_categories}. 
If a benchmark does not come with a validation split, we randomly allocate 20\% of the training data as the validation set.  
To save computational resources, we cap the number of training data at 50,000, validation data at 1,000, and testing data at 10,000.

\ifbool{show_model_table}{
\begin{table}[t]
    \caption{Models considered, categorized by model family.}
    \label{app_tab:models}
    \centering
    \resizebox{1.0\textwidth}{!}{
    \begin{tabular}{c l}
    \toprule
    Family & Model Name Suffix \\
    \midrule
    \textsc{Llama-} & \makecell[l]{
    \textsc{3-8B}, \textsc{3.1-8B},
    \textsc{3.2-1B}, \textsc{3.2-3B}, 
    \textsc{3-8B-IT}, \textsc{3.1-8B-IT}, \textsc{3.2-1B-IT}, \textsc{3.2-3B-IT}
    } \\[0.4em]
    % \hline
    \textsc{Qwen-}  & \makecell[l]{\textsc{1.5-0.5B}, \textsc{1.5-1.8B}, \textsc{1.5-4B}, \textsc{1.5-7B}, \textsc{1.5-14B}, \textsc{2-0.5B}, \textsc{2-1.5B}, \textsc{2-7B},  \textsc{2.5-0.5B}, \textsc{2.5-1.5B}, 
    \textsc{2.5-3B}, \textsc{2.5-7B},\\ \textsc{2.5-14B},  \textsc{1.5-0.5B-IT}, \textsc{1.5-1.8B-IT}, \textsc{1.5-4B-IT},  \textsc{1.5-7B-IT}, \textsc{1.5-14B-IT},  
    \textsc{2-0.5B-IT}, \textsc{2-1.5B-IT}, \textsc{2-7B-IT},\\  \textsc{2.5-0.5B-IT}, \textsc{2.5-1.5B-IT}, \textsc{2.5-3B-IT},  \textsc{2.5-7B-IT}, \textsc{2.5-14B-IT}
    }\\[0.8em]
    % \hline
    \textsc{Gemma-} & \makecell[l]{\textsc{2B}, \textsc{7B}, \textsc{2-2B}, \textsc{2-9B}, \textsc{2B-IT}, \textsc{7B-IT}, \textsc{2-2B-IT}, \textsc{2-9B-IT}} \\[0.4em]
    % \hline
    \textsc{GPT2-} & \makecell[l]{\textsc{124M}, \textsc{335M}, \textsc{774M}, \textsc{1.5B}}\\[0.4em]
    % \hline
    \textsc{Pythia-} & \makecell[l]{\textsc{70M}, \textsc{160M}, \textsc{410M}, \textsc{1B}, \textsc{1.4B}, \textsc{2.8B}, \textsc{6.9B}, \textsc{12B}} \\[0.4em]
    % \hline
    \textsc{Yi-} & \makecell[l]{\textsc{6B}, \textsc{9B}, \textsc{6B-IT}, \textsc{1.5-6B}, \textsc{1.5-9B}, \textsc{1.5-6B-IT}, \textsc{1.5-9B-IT}}  \\[0.4em]
    \bottomrule
    \end{tabular}}
\end{table}}{}

\paragraph{Model selection.} 
We consider 61 language models across six model families: \textsc{Llama}~\citep{llama}, \textsc{Qwen}~\citep{qwen}, \textsc{Gemma}~\citep{gemma}, \textsc{Pythia}~\citep{pythia}, \textsc{GPT-2}~\citep{gpt2} and \textsc{Yi}~\citep{yi}. Due to computational constraints, we select models with no more than 14B parameters. See Table~\ref{app_tab:models} \ifbool{show_model_table}{}{in the Appendix~\ref{app:model_setting}} for the full list. We include both base and instruction-tuned models, and use the suffix \textsc{-IT} to denote instruction-tuned models.

\paragraph{Evaluation setup.} 
We evaluate the 61 models across all 24 benchmarks using both direct evaluation and train-before-test evaluation. 
We use the \texttt{lm-eval-harness} library for evaluation. 
We evaluate models zero-shot~\citep{Brown2020LanguageMA}. 
For direct evaluation, we simply evaluate the model as it is.
For train-before-test, we fine-tune models for five epochs using learning rates in $\{1\mathrm{e}-5, 2\mathrm{e}-5, 5\mathrm{e}-5\}$, separately.
The best performing checkpoint is then selected based on performance on a separate validation set, yielding $61\times24=1,464$ fine-tuned models in total.
We use parameter-efficient fine-tuning (PEFT)~\citep{Hu2021LoRALA,peft}.
See more details in Appendix~\ref{app:eval_setting}.
Each fine-tuned model is then evaluated on the corresponding benchmark's test set. For each benchmark, we rank models according to their performance. We then measure the ranking correlation across pairs of benchmarks using Kendall’s $\tau$~\citep{kendall1938new}.

\begin{figure}[t]
    \centering
    \includegraphics[width=0.99\linewidth]{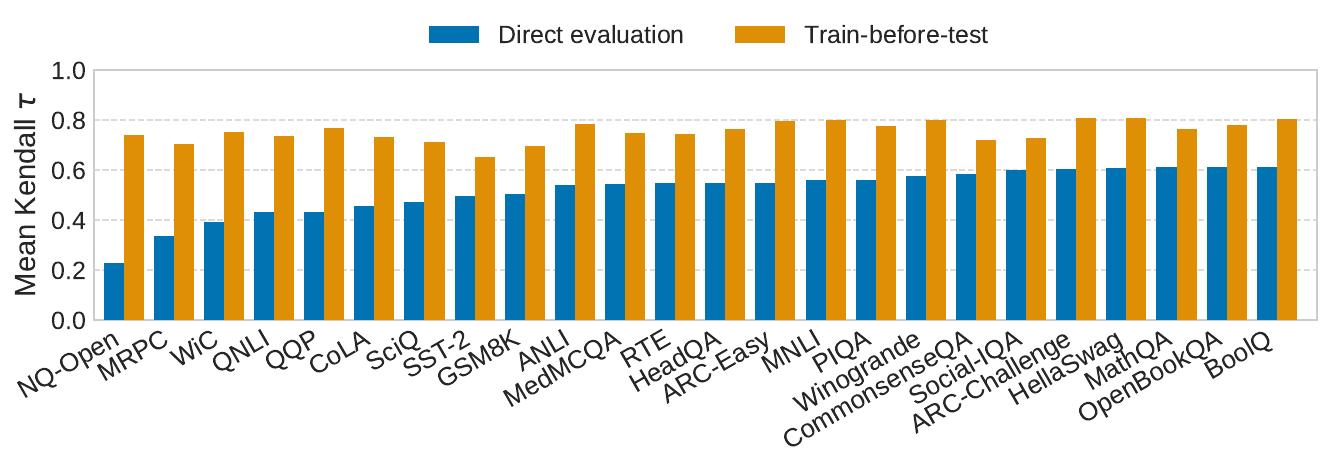}
    \caption{
    Mean ranking agreement between each benchmark and all others. We calculate Kendall’s~$\tau$ between each benchmark and every other benchmark, and then average the results. Compared to direct evaluation, train-before-test consistently improves ranking agreement, often by a large margin. A detailed comparison of Kendall's $\tau$ values for every benchmark pair is provided in Appendix~\ref{app:cross_task_agree}. On average, the overall average Kendall’s $\tau$ is 0.52 for direct evaluation and 0.76 for train-before-test.
    }
    \label{fig:downstream_consistency}
\end{figure}

\begin{figure}[t]
\centering
\begin{subfigure}{
\ifbool{show_full_size_fig}{0.45}{0.4}\textwidth
}
  \centering
  \includegraphics[width=\linewidth]{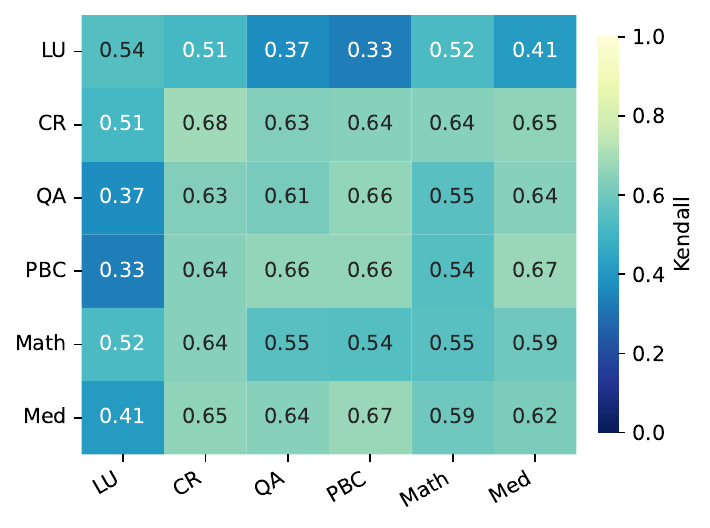}
  \caption{Direct evaluation.}
\end{subfigure}%
\begin{subfigure}{
\ifbool{show_full_size_fig}{0.45}{0.4}\textwidth
}
  \centering
  \includegraphics[width=\linewidth]{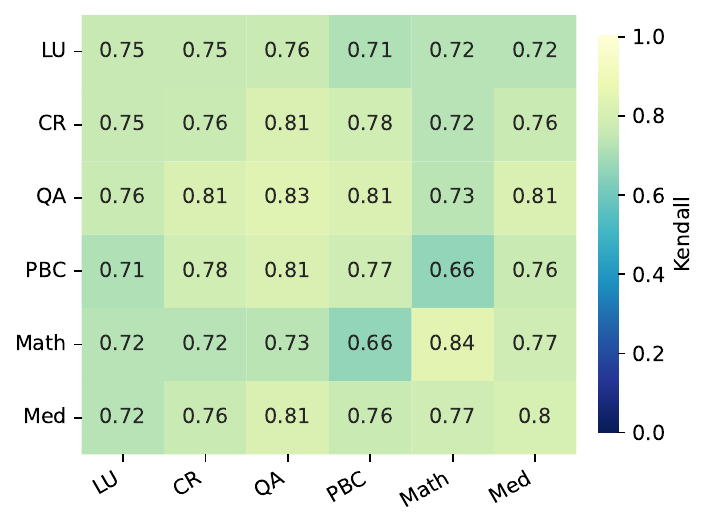}
  \caption{Train-before-test.}
\end{subfigure}
\caption{
Cross-category ranking agreement for direct evaluation (left) and train-before-test (right). 
We categorize benchmarks into language understanding (LU), commonsense reasoning (CR), question answering (QA), physics/biology/chemistry (PBC), math (Math), and medicine (Med), see Table~\ref{tab:benchmark_categories}. 
Kendall’s $\tau$ is averaged across all pairs of benchmarks that belong to two given categories.
The diagonal entries represent intra-category agreement and the others represent inter-category agreement.
Train-before-test improves both intra- and inter-category ranking agreement in all instances.
}
\label{fig:cross_category_consistency}
\end{figure}

\subsection{Downstream Ranking Agreement}

As depicted in Figure~\ref{fig:downstream_consistency}, direct evaluation shows only modest ranking agreement between the 24 benchmarks, with an average Kendall's $\tau$ of 0.52.
This lack of agreement across benchmarks complicates model assessment and makes it challenging to aggregate results into a meaningful overall ranking~\citep{zhang2024inherent}. 
In contrast, the train-before-test methodology leads to a substantial improvement in ranking agreement. 
Under this approach, 274 out of 276 benchmark pairs show higher Kendall’s $\tau$ scores, with the average $\tau$ rising from 0.52 to 0.76.
This stronger consistency suggests that model potential ranking on one benchmark is likely to generalize to others, including practitioners’ own cases, which simplifies model comparison and selection.
Notably, benchmarks that appeared to be outliers under direct evaluation, such as \texttt{NQ-Open} and \texttt{MRPC}, demonstrate much greater ranking consistency under train-before-test.
For example, the average Kendall's $\tau$ between \texttt{NQ-Open} and all other benchmarks improves from 0.23 to 0.74.

We further split all benchmarks into six categories (e.g., language understanding, math), see Table~\ref{tab:benchmark_categories}.
For each category pair, we report in Figure~\ref{fig:cross_category_consistency} the intra-category average ranking correlations and inter-category average ranking correlations across all relevant benchmark pairs. 
Consistent with our previous findings, we observe reasonably poor ranking agreements across categories under direct evaluation. 
While one might expect high intra-category agreement---after all, tasks within the same category tend to be relatively similar---direct evaluation results in low intra-category agreement in many cases. 
For example, the intra-category mean Kendall's $\tau$ is 0.54 for language understanding and 0.55 for math. 
This further underscores the difficulty of selecting models based on direct evaluation. 
Even if the goal is to choose a model that excels within a specific domain, the low intra-category agreement makes this decision challenging.

In contrast, train-before-test boosts both intra- and inter-category consistency. 
For example, the intra-category mean Kendall's $\tau$ for language understanding raises from 0.52 to 0.75, as well as from 0.55 to 0.84 for the math category. 
Moreover, agreement between categories is often nearly as high as agreement within categories.
This suggests that models with higher potential in one domain tend to also perform well across other domains after adaptation.

\subsection{Perplexity Agreement}
We now study the agreement between downstream benchmark rankings and perplexity rankings on general domain corpora. 
To do so, we collect three corpora from \texttt{Wiki}pedia, \texttt{Stack}Exchange, and \texttt{arXiv}, retaining only contents from 2025. 
Because all models used were released before 2025, they could not have seen these texts during pretraining.
Specifically,  we collect 3,366 documents for \texttt{Wiki}, 6,001 for \texttt{Stack}Exchange and 44,384 for \texttt{arXiv}. 
These datasets are split into training, validation, and testing sets, in an 8:1:1 ratio.

\begin{figure}[t]
\centering
\begin{subfigure}{\textwidth}
  \centering
  \includegraphics[width=\linewidth]{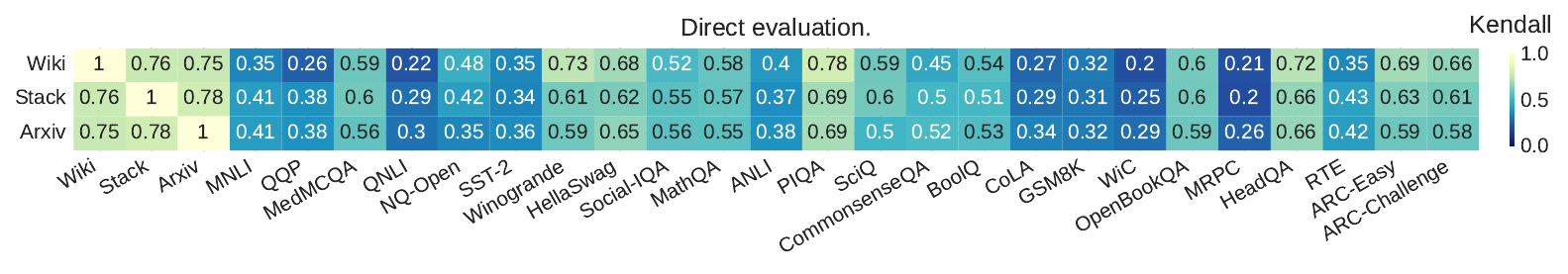}
\end{subfigure}\\
\begin{subfigure}{\textwidth}
  \centering
  \includegraphics[width=\linewidth]{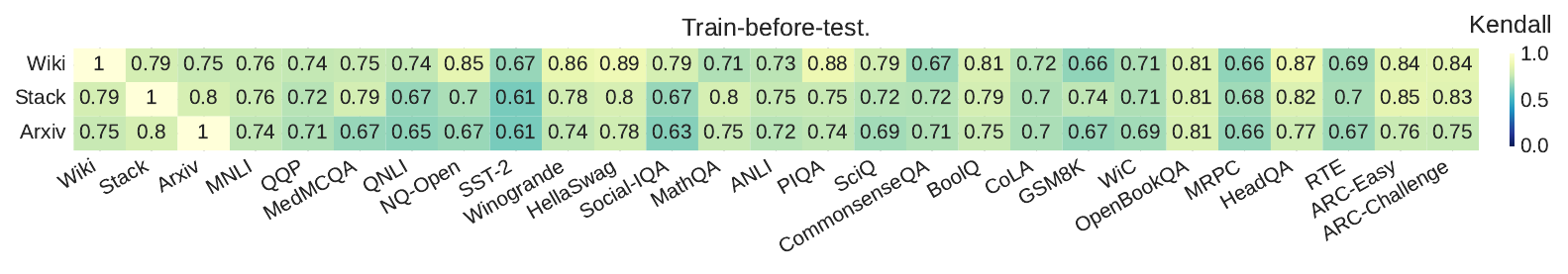}
\end{subfigure}
\caption{Ranking agreement between perplexity rankings and downstream benchmark rankings under direct evaluation (top) and train-before-test (bottom).  
Perplexity rankings are consistent with each other under both evaluation schemes, with an average Kendall's $\tau$ of 0.76 and 0.78, respectively. However, for direct evaluation, agreement between perplexity rankings and downstream rankings is low, with an average Kendall's $\tau$ of just 0.48. Fortunately, train-before-test results in much higher agreement between perplexity and downstream evaluations, increasing average Kendall's $\tau$ to 0.74. 
}
\label{fig:pp}
\end{figure}

\begin{figure}[t]
\centering
\begin{subfigure}{\textwidth}
  \centering
  \includegraphics[width=\linewidth]{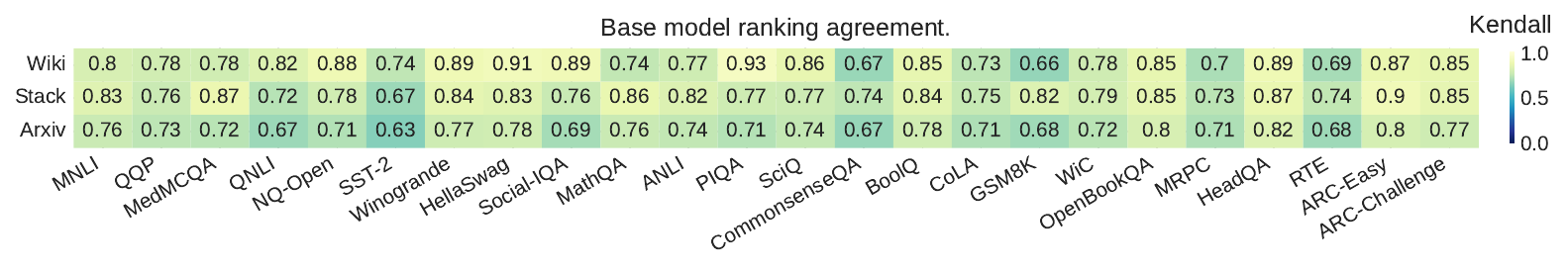}
\end{subfigure}\\
\begin{subfigure}{\textwidth}
  \centering
  \includegraphics[width=\linewidth]{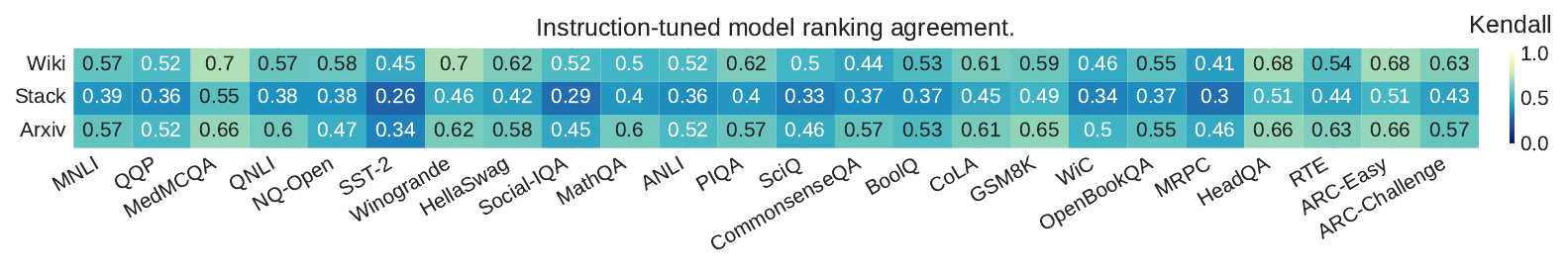}
\end{subfigure}
\caption{
Ranking agreement between perplexity rankings \textbf{before fine-tuning} (direct evaluation) and downstream benchmark rankings \textbf{after fine-tuning} (train-before-test) for base models (top) and instruction-tuned models (bottom).
Unlike Figure~\ref{fig:pp} where both rankings in each comparison use the same evaluation scheme, here we test whether pre-fine-tuning perplexity can predict post-fine-tuning downstream performance.
Base models show strong correlation (average Kendall's $\tau$ = 0.78), suggesting perplexity is a good predictor of model potential. 
Instruction-tuned models show much weaker correlation (average Kendall's $\tau$ = 0.51).
}
\label{fig:pp_dme}
\end{figure}

We measure perplexity in bits per byte, using the \texttt{lm-eval-harness} library. We then compute models rankings based on the perplexity evaluations, and compare the rankings with those of the downstream benchmarks considered in earlier sections. We exclude the four \textsc{Gemma} models from these results, as \texttt{lm-eval-harness} provides unreliable perplexity measurements for \textsc{Gemma} models due to its rolling window implementation. See Appendix~\ref{app:perplexity} for details.

The results are presented in Figure~\ref{fig:pp}. 
In contrast to downstream tasks, perplexity rankings demonstrate strong agreement both under direct evaluation and train-before-test. 
Specifically, the average Kendall’s~$\tau$ between the perplexity rankings is 0.76 for direct evaluation and 0.78 for train-before-test. We hypothesize that this reasonably strong agreement arises due to the smooth relationship between perplexity evaluations~\citep{brandfonbrener2024loss, mayilvahananllms}.

When comparing ranking agreement between perplexity evaluations and downstream benchmarks, we find that agreement is low under direct evaluation, with a mean Kendall’s $\tau$ of 0.48. 
This lack of agreement is concerning, as it signals a disconnect between the language modelling pre-training objective and downstream benchmark performance. 
Fortunately, we find that our train-before-test methodology improves ranking agreement substantially, with the mean Kendall’s $\tau$ ranking correlation between perplexity rankings and benchmark rankings rising to 0.74. 
This finding is reassuring: a light amount of fine-tuning on task data is sufficient to align the language modeling training objective with downstream performance. 
Moreover, we find that ranking agreement between perplexity and downstream evaluations is roughly similar to agreement across downstream evaluations. 
This suggests that, despite perplexity typically not being used for benchmarking purposes, it can be as effective a ranking metric as benchmark evaluations.

Drawing inspiration from prior work~\citep{liu2023same, xia2023training, Gadre2024LanguageMS, benchmarkloss,Zhang2024WhenSM}, we further examine the correlation between model rankings according to \emph{average} perplexity across the three text corpora and \emph{average} downstream performance across the 24 benchmarks. 
\ifbool{use_citet}{\citet{Gadre2024LanguageMS}}{Zhang and Hardt~(2024, \citep{Gadre2024LanguageMS})} show that, when models are trained on the same pretraining data, perplexity is well-correlated with aggregate benchmark performance. 
Our setup differs in that we consider a diverse set of model families, each trained on different pretraining data. 
Under direct evaluation, we find that the ranking correlation is modest, with a Kendall’s $\tau$ of only 0.55. 
We hypothesize that this relatively weak agreement is due to differences in pretraining data and instruction tuning, resulting in varying levels of exposure to benchmark tasks during training~\citep{dominguez2024training}.
Fortunately, when applying our train-before-test methodology, the ranking correlation between average perplexity and average downstream performance improves substantially, with Kendall’s $\tau$ increasing from 0.55 to 0.84.

We additionally examine the agreement between perplexity prior to fine-tuning and downstream task performance after fine-tuning. 
That is, between direct evaluation perplexity rankings and train-before-test downstream performance rankings. 
We plot such ranking agreement in Figure~\ref{fig:pp_dme}, dividing models into base models and instruction-tuned models. 
For base models, perplexity prior to fine-tuning is a strong indicator of model potential on downstream tasks, with an average Kendall's $\tau$ of 0.78. 
This indicates that, for base models, direct evaluation of perplexity is already a reasonably reliable metric for ranking models. 
Moreover, it indicates that the ranking consistency we observe reflects inherent model potential rather than artifacts introduced by fine-tuning.

However, the same does not hold for instruction-tuned models (average Kendall's $\tau$ = 0.51).
Instruction-tuning renders perplexity rankings unreliable, as ranking agreement is low across the board. 
This is to be expected: instruction fine-tuning tends to increase both benchmark performance ($\uparrow$) and perplexity ($\downarrow$) on general text corpora, thus clouding the relationship between perplexity and downstream evaluations. 
Fortunately, as shown earlier, train-before-test restores high ranking agreement between perplexity evaluations and downstream performance.

\begin{figure}[t]
    \centering
    \includegraphics[width=\ifbool{show_full_size_fig}{0.45}{0.4}\linewidth]{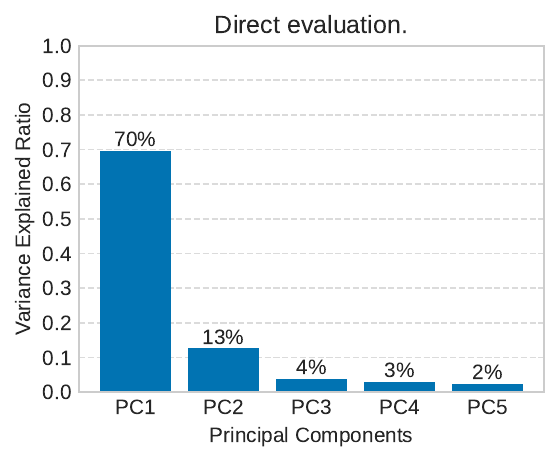}
    \includegraphics[width=\ifbool{show_full_size_fig}{0.45}{0.4}\linewidth]{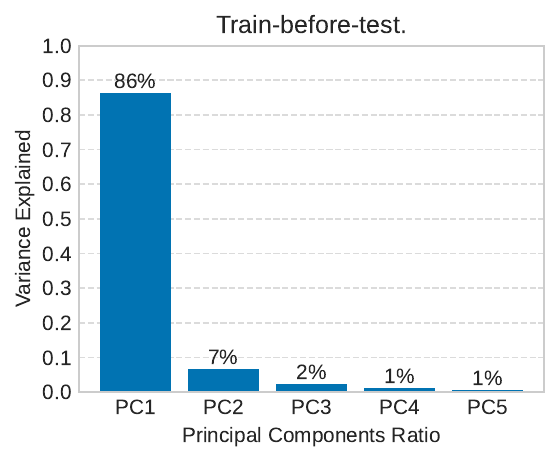}
    \caption{
    Explained variance ratios of the top five principal components of the benchmark score matrix, under direct evaluation (left) and train-before-test (right).
    Train-before-test substantially increases the amount of variance explained by the first principal component, from 70\% to 86\%
    }
    \label{fig:pca}
\end{figure}

\subsection{
Low-Ranked Model Score Matrix
}
\label{sec:pca}

So far, we have shown that comparing model potential using the train-before-test yields consistent rankings across benchmarks.
We now examine the implications of this finding by analyzing the resulting matrix of model scores, where each entry $(i, j)$ corresponds to the performance of model~$j$ on a benchmark~$i$. We use Principal Component Analysis (PCA) to examine the structure of the model score matrix.

\begin{wrapfigure}{r}{0.43\textwidth}
    \vspace{-6mm}
    \centering
    \includegraphics[width=1.0\linewidth]{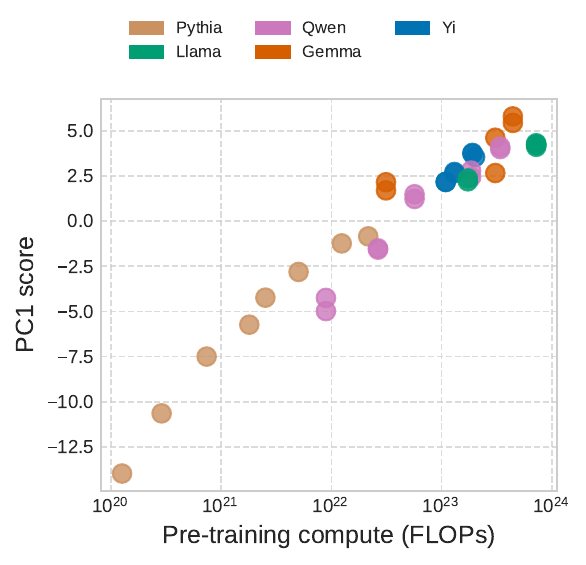}
    \caption{
    PC1 scores under train-before-test align with the pre-training compute.}
    \label{fig:pc1_scale}
    \vspace{-4mm}
\end{wrapfigure}
Figure~\ref{fig:pca} shows the explained variance ratios of the first five principal components. 
These results support previous findings that the score matrix is of low rank~\citep {Ruan2024ObservationalSL}. 
Under direct evaluation, the first five components account for 91\% of the total variance. 
A similar trend is observed for train-before-test scores, where the first five components explain 97\% of the variance. 
Notably, under train-before-test, the first principal component (PC1) captures a much larger share of the variance: 86\%, compared to 70\% for direct evaluation. 

Prior works interpret PC1 scores under direct evaluation as an indication of general capability, with later principal components denoting domain-specific capabilities not captured by PC1 (e.g., reasoning ability, coding ability)~\citep{Ruan2024ObservationalSL,Burnell2023RevealingTS}.
Unlike out-of-the-box performance, which is controlled by multiple factors~\citep{Ruan2024ObservationalSL,Burnell2023RevealingTS}, model potential is dominated by one single principal axis.
It is of no surprise that PC1 also positively correlates with pre-training compute, as shown in Figure~\ref{fig:pc1_scale}\footnote{We only plot models whose number of training tokens is publicly available. See Table~\ref{tab:flops} for details.}, which have been identified as crucial to model performances~\citep{kaplan2020scaling,Hoffmann2022TrainingCL}.
See detailed PC1 scores in Figure~\ref{app_fig:pc1_score}.

\paragraph{Case study for Qwen models.}
\begin{figure}[t]
    \centering
    \includegraphics[width=\ifbool{show_full_size_fig}{0.45}{0.4}\linewidth]{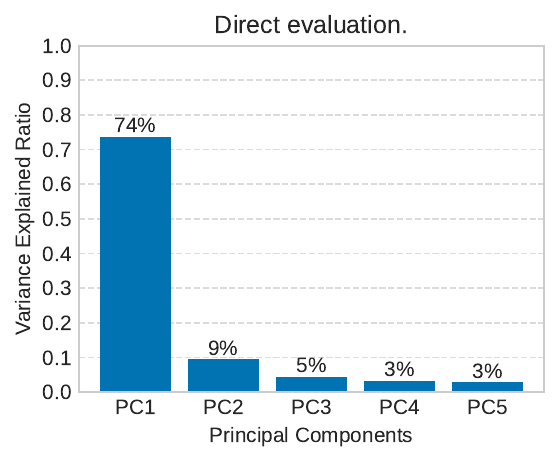}
    \includegraphics[width=\ifbool{show_full_size_fig}{0.45}{0.4}\linewidth]{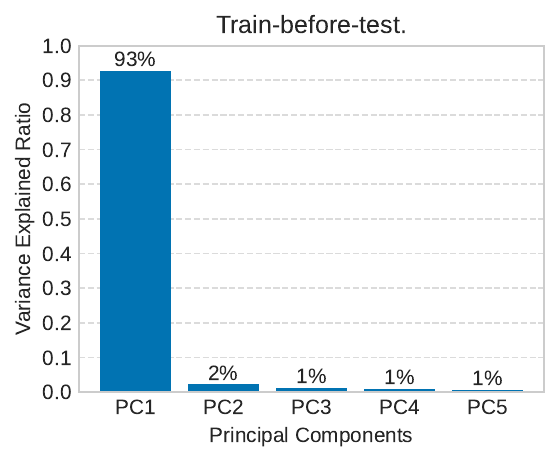}
    \caption{
    Explained variance ratios of the top five principal components of the \textsc{Qwen}  score matrix. For train-before-test, the explained variance ratio of PC1 increases to 93\%, making the \textsc{Qwen} score matrix essentially rank one.
    }
    \label{fig:pca_qwen}
\end{figure}

We repeat our PCA analysis on the score matrix containing only \textsc{Qwen} models, depicted in Figure~\ref{fig:pca_qwen}. Remarkably, we find that PC1 for train-before-test explains 93\% of the variance, roughly as much as the variance explained by the top five principal components under direct evaluation. That is, whereas for direct evaluation the score matrix is low-rank; train-before-test renders the score matrix essentially rank one.

%% file: sections/discussion.tex
\section{Discussion, limitations, and conclusion}

Train-before-test fundamentally reframes how we interpret model evaluation. Whereas direct evaluation yields benchmark-specific rankings that often contradict one another, train-before-test harmonizes rankings across a wide array of tasks and datasets. This shift from measuring out-of-the-box \emph{performance} to comparing achievable \emph{potential} equips the community with a more stable and externally valid evaluation methodology.

This emphasis on model potential is particularly valuable for scenarios involving model development and adaptation. 
Practitioners frequently need to make decisions during model development---selecting checkpoints mid–pre-training or choosing a base model for further instruction tuning or domain-specific adaptation.
In these scenarios, direct evaluation, while useful for assessing deployment readiness, is of limited relevance and utility.
A model that performs poorly on direct evaluation might excel when adapted to new tasks. 
Train-before-test, by contrast, shows that rankings on any task will also generalize to others, offering more promising guidance for model selection.

One might argue that ranking consistency is unnecessary if we can simply choose benchmarks close to a given downstream application. However, our findings highlight three challenges with that view. First, even benchmarks that purport to measure the same skill (\emph{e.g.}, question answering) produce contradictory rankings under direct evaluation. Second, no benchmark perfectly captures the specifics of an application, making some degree of generalization unavoidable. Third, in real deployments, models are often adapted to varying degrees, making the \emph{potential} the relevant signal for comparison. 

\textbf{Limitations.}
Train-before-test requires fine-tuning models on task-specific data before evaluation, which certainly increases the evaluation cost. However, this investment yields dividends through improved reliability. 
Our findings suggest that fewer benchmarks suffice under train-before-test, as rankings from one benchmark reliably transfer to others. 
This reduction in required evaluations can offset the per-benchmark cost increase.
A second problem is that, unfortunately, many benchmarks no longer come with training data, making it more difficult to apply train-before-test. In light of our findings, we recommend that future benchmarks provide fine-tuning data for the benchmark. A third limitation is that some commercial model providers do not easily allow fine-tuning of their models. We contend that in this case the problem is with the model provider. There is clearly scientific value in creating an ecosystem of models that can be fine-tuned. Train-before-test evaluation creates additional incentives for making models easy to fine-tune.

\textbf{Conclusion.}
Overall, train-before-test complements existing evaluation practices by distinguishing between \emph{performance} and \emph{potential}. 
cImportantly, potential comparison is not intended to replace direct evaluation---both serve distinct purposes. 
Direct evaluation remains useful for understanding immediate deployment readiness, while potential comparison provides insights into adaptability and development prospects. 
Together, they offer a more complete picture of model capabilities. 
We believe that adopting train-before-test as a standard alongside direct evaluation can significantly improve the reliability, interpretability, and practical utility of the model evaluation ecosystem.

%% file: sections/ackownledge.tex
\section*{Acknowledgement}
We thank Yatong Chen, Florian Dorner, Mina Remeli and Jiduan Wu for helpful discussions and feedback on draft versions of this work.

%% file: sections/appendix.tex
\appendix
\onecolumn
\section{Additional Experiment Setting}
\label{app:setting}
\subsection{Benchmark Selection}
\label{app:data_setting}

We begin our study with the \texttt{lm-eval-harness} package~\citep{eval-harness}, which offers a comprehensive suite of language model benchmarks. 
To accommodate the train-before-test methodology which requires a dedicated training set for fine-tuning, we first identify benchmarks that provide at least 1,000 training examples. 
This yields a total of 37 benchmarks. 

These benchmarks can be broadly categorized as 28 likelihood-based and 9 generation-based benchmarks.
Likelihood-based evaluations test for the likelihood of different completions given some input string; for example, different answer choices given a multiple-choice input question. 
Since the number of completions is usually small, likelihood-based evaluations are generally compute-efficient. 

Generation-based evaluations, in contrast, generate some output response given an input query. If responses tend to be long, then generation-based evaluations naturally become compute-intensive. 
This is particularly true for base models, which are usually not trained for instruction following and therefore continue to generate tokens until hitting their maximum token limit. 
These generation-based benchmarks are also over-challenging for smaller models with limited parameters, such as \textsc{GPT-2}~\citep{gpt2}.
Therefore, we exclude seven generation-based benchmarks, \texttt{Drop}, \texttt{CoQa}, \texttt{ReCoRD}, \texttt{bAbi}, \texttt{WebQA}, \texttt{TriviaQA} and \texttt{Fld-Default}. 
Nevertheless, we retain two widely used generation-based benchmarks, \texttt{GSM8K} and \texttt{NQ-Open}, in our experiments.

We additionally excluded five benchmarks due to anomalies observed during fine-tuning: \texttt{MedQA-4Options}, \texttt{LogiQA}, \texttt{Mutual}, \texttt{Mela-EN}, and \texttt{Swag}. 
For these benchmarks, more than 20\% of models, most of which are small models with fewer than 0.5B parameters, showed no performance improvement after fine-tuning.
We also excluded \texttt{Paws-EN}, as its corresponding model ranking under direct evaluation was negatively correlated (Kendall’s $\tau$ less than zero) with 23 out of 24 other benchmarks. We attribute this anomaly to the unusual prompting template used by \texttt{lm-eval-harness}.

Our final selection includes 24 benchmarks:
\texttt{MNLI}~\citep{mnli}, \texttt{QNLI}~\citep{qnli}, \texttt{RTE}~\citep{rte,rte2,rte3}, \texttt{CoLA}~\citep{cola}, \texttt{SST-2}~\citep{sst2}, \texttt{MRPC}~\citep{mrpc}, \texttt{QQP}, \texttt{WiC}~\citep{wic}, \texttt{ANLI}~\citep{anli},
\texttt{Winogrande}~\citep{winograd}, \texttt{CommonsenseQA}~\citep{commonsenseqa},  \texttt{Hellaswag}~\citep{hellaswag}, \texttt{Social-IQA}~\citep{socialiqa},
\texttt{OpenBookQA}~\citep{openbookqa}, \texttt{NQ-Open}~\citep{nq_open}, \texttt{BoolQ}~\citep{boolq}, \texttt{ARC-Easy}, \texttt{ARC-Challenge}~\citep{allenai:arc},
\texttt{SciQ}~\citep{sciq}, \texttt{PIQA}~\citep{piqa},
\texttt{MathQA}~\citep{mathqa}, \texttt{GSM8K}~\citep{gsm8k},
 \texttt{MedMCQA}~\citep{medmcqa}, \texttt{HeadQA}~\citep{headqa}.

\ifbool{show_model_table}{}{
\subsection{Model Selection}
\label{app:model_setting}
See Table~\ref{app_tab:models} for the complete list of models used in our experiments.

\begin{table}[t]
    \caption{Models considered, categorized by model family.
    % We mark each model by its generation within the family and number of model parameters and use the suffix \textsc{-IT} to denote instructed models.
    }
    \label{app_tab:models}
    \centering
    \resizebox{1.0\textwidth}{!}{
    \begin{tabular}{c l}
    \toprule
    Family & Model Name Suffix \\
    \midrule
    \textsc{Llama-}~\citep{llama} & \makecell[l]{
    \textsc{3-8B}, \textsc{3.1-8B},
    \textsc{3.2-1B}, \textsc{3.2-3B}, 
    \textsc{3-8B-IT}, \textsc{3.1-8B-IT}, \textsc{3.2-1B-IT}, \textsc{3.2-3B-IT}
    } \\[0.4em]
    % \hline
    \textsc{Qwen-}~\citep{qwen}  & \makecell[l]{\textsc{1.5-0.5B}, \textsc{1.5-1.8B}, \textsc{1.5-4B}, \textsc{1.5-7B}, \textsc{1.5-14B}, \textsc{2-0.5B}, \textsc{2-1.5B}, \textsc{2-7B},  \textsc{2.5-0.5B}, \textsc{2.5-1.5B}, \\ 
    \textsc{2.5-3B}, \textsc{2.5-7B}, \textsc{2.5-14B},  \textsc{1.5-0.5B-IT}, \textsc{1.5-1.8B-IT}, \textsc{1.5-4B-IT},  \textsc{1.5-7B-IT}, \textsc{1.5-14B-IT},\\  
    \textsc{2-0.5B-IT}, \textsc{2-1.5B-IT}, \textsc{2-7B-IT}, \textsc{2.5-0.5B-IT}, \textsc{2.5-1.5B-IT}, \textsc{2.5-3B-IT},  \textsc{2.5-7B-IT}, \textsc{2.5-14B-IT}
    }\\[0.8em]
    % \hline
    \textsc{Gemma-}~\citep{gemma} & \makecell[l]{\textsc{2B}, \textsc{7B}, \textsc{2-2B}, \textsc{2-9B}, \textsc{2B-IT}, \textsc{7B-IT}, \textsc{2-2B-IT}, \textsc{2-9B-IT}} \\[0.4em]
    % \hline
    \textsc{GPT2-}~\citep{gpt2} & \makecell[l]{\textsc{124M}, \textsc{335M}, \textsc{774M}, \textsc{1.5B}}\\[0.4em]
    % \hline
    \textsc{Pythia-}~\citep{pythia} & \makecell[l]{\textsc{70M}, \textsc{160M}, \textsc{410M}, \textsc{1B}, \textsc{1.4B}, \textsc{2.8B}, \textsc{6.9B}, \textsc{12B}} \\[0.4em]
    % \hline
    \textsc{Yi-}~\citep{yi} & \makecell[l]{\textsc{6B}, \textsc{9B}, \textsc{6B-IT}, \textsc{1.5-6B}, \textsc{1.5-9B}, \textsc{1.5-6B-IT}, \textsc{1.5-9B-IT}}  \\[0.4em]
    \bottomrule
    \end{tabular}}
\end{table}}

\subsection{Evaluation Setup}
\label{app:eval_setting}
For our train-before-test evaluations, we fine-tune each model for five epochs and select the best-performing checkpoint based on evaluations on a separate validation set. 
We use the AdamW optimizer with a weight decay of 0.01. 
For each model-benchmark combination, we perform a hyperparameter search over three learning rates $\{1e-5, 2e-5, 5e-5\}$ and select the optimal one based on validation performance. 
To reduce memory consumption, we employ parameter-efficient fine-tuning (PEFT)~\citep{Hu2021LoRALA,peft}, 
We use a LoRA configuration with rank 8, $\alpha=32$, and dropout 0.1.
Most of our experiments are conducted on Quadro RTX 6000, Tesla V100-SXM2-32GB and NVIDIA A100-SXM4-80GB GPUs.

In cases where models show no performance improvement after fine-tuning, we report their pre-fine-tuning results. 
This scenario is rare and typically occur with smaller models (less than 500M parameters) that lack the capacity to perform certain tasks, resulting in near-random performance both before and after fine-tuning.
Additionally, since all training datasets in our study are publicly available, some models may have encountered this data during pre-training, potentially limiting the benefits of additional fine-tuning.

For instruction-tuned models, we evaluate performance both with and without chat templates, selecting the configuration that yields better results. 
Specifically, during direct evaluation, we assess model performance on the validation set under both conditions and apply the better-performing configuration to the test set. 
In the train-before-test setting, we similarly fine-tune two variants: one with training data formatted using chat templates and one without. 
We then select the approach that achieves the best performance on the validation set for final evaluation.

\clearpage
\section{Additional Experiment Results}
\label{app:results}
\subsection{Downstream Ranking Agreement}
\label{app:cross_task_agree}

\begin{figure}[t]
    \centering
    \includegraphics[width=0.99\linewidth]{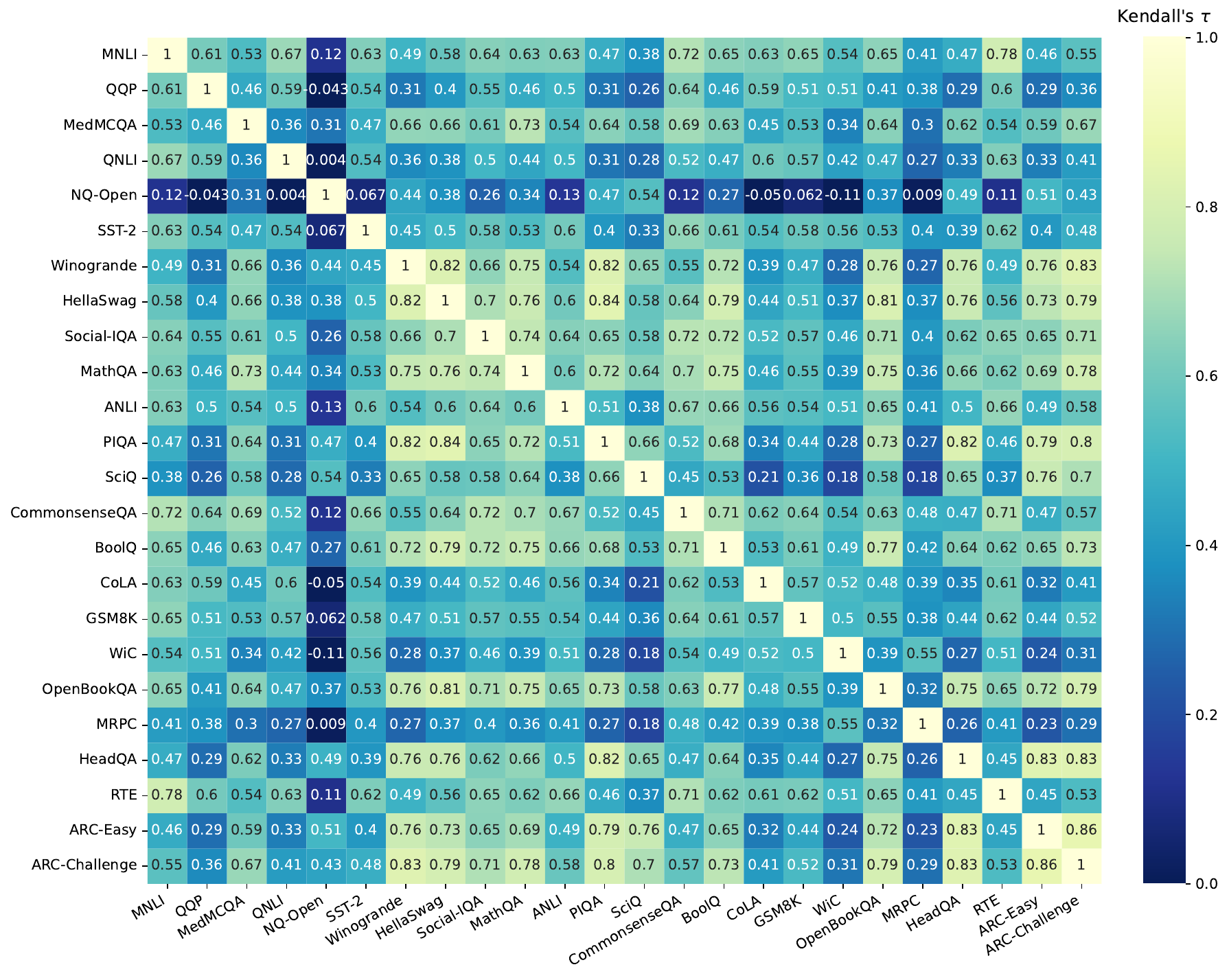}
    \caption{
    Cross benchmark ranking agreement under direct evaluation. 
    Benchmarks are sorted based on the training dataset size.
    Kendall’s $\tau$ is calculated for every benchmark pair. 
    }
    \label{app_fig:detail_direct_eval}
\end{figure}

\begin{figure}[t]
    \centering
    \includegraphics[width=0.99\linewidth]{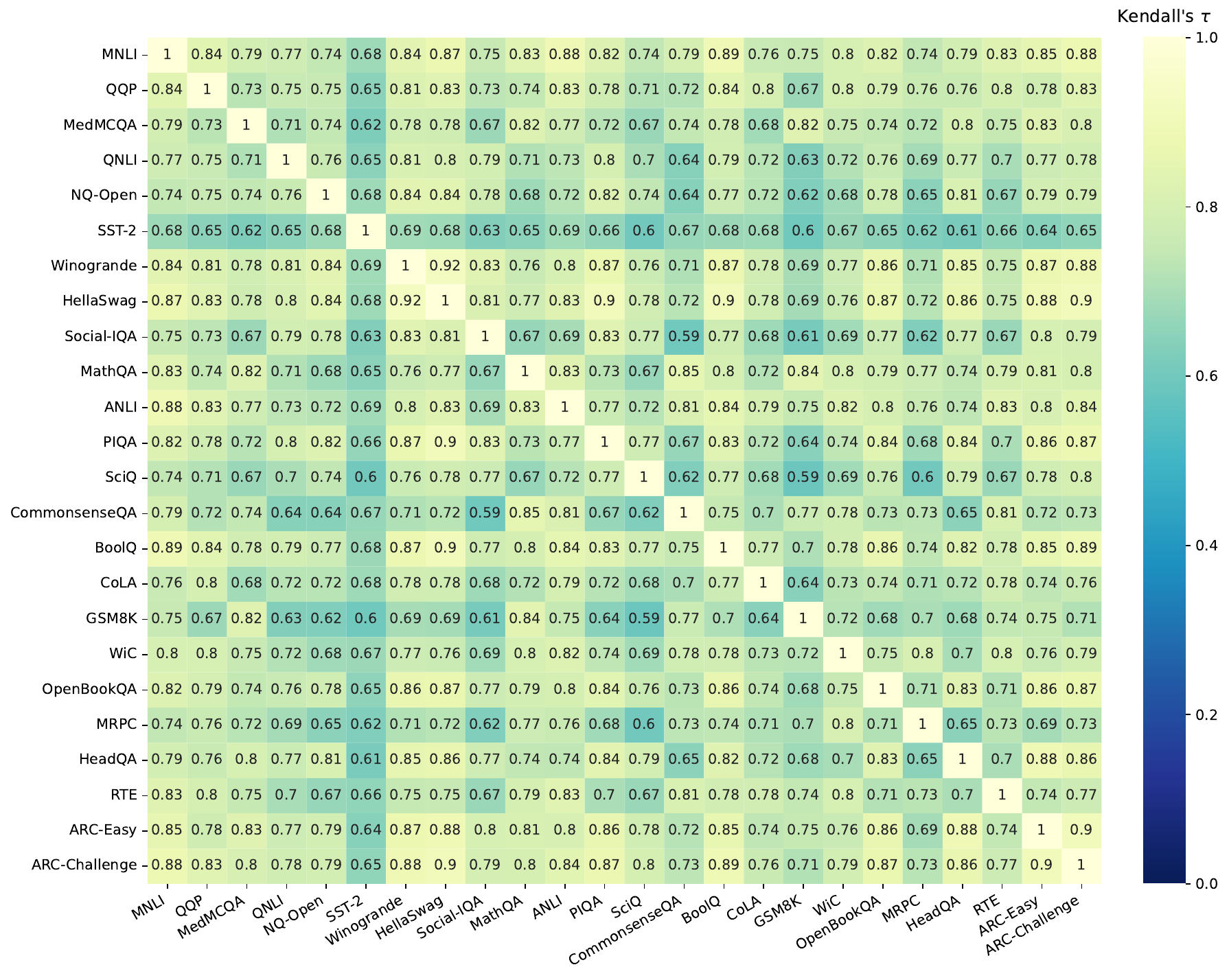}
    \caption{
    Cross benchmark ranking agreement under train-before-test.
    Benchmarks are sorted based on the training dataset size.
    Kendall’s $\tau$ is calculated for every benchmark pair. 
    }
    \label{app_fig:detail_train_before_test}
\end{figure}

We plot detailed pairwise ranking correlation agreement between benchmarks in Figure~\ref{app_fig:detail_direct_eval} (direct evaluation) and \ref{app_fig:detail_train_before_test} (train-before-test), corresponding to Figure~\ref{fig:downstream_consistency} in the main text.

\clearpage
\subsection{Perplexity Ranking Agreement}
\label{app:perplexity}
\begin{table}[t]
    \centering
    \caption{
    Bits per byte (BPB) of eight excluded \textsc{Gemma} models compared to \textsc{Pythia-410M} across the three newly collected corpora. 
    The \textsc{Gemma} models exhibit abnormally high BPB values on \texttt{Wiki} and \texttt{Stack}, likely due to the greater average sequence length in these two datasets. 
     Specifically, \texttt{Arxiv} has an average of 163 words per document, compared to 250 for \texttt{Stack} and 1502 for \texttt{Wiki}.
    }
    \label{app_tab:perplexity}
    \begin{tabular}{llll}
    \toprule
     & \texttt{Arxiv} & \texttt{Wiki} & \texttt{Stack} \\
    \midrule
    \textsc{Gemma-2B} & 0.766 & 1.578 & 1.139 \\
    \textsc{Gemma-2B-IT} & 0.770 & 1.524 & 1.222 \\
    \textsc{Gemma-7B} & 1.013 & 4.780 & 4.053 \\
    \textsc{Gemma-7B-IT} & 1.053 & 18.711 & 20.958 \\
    \textsc{Gemma-2-2B} & 0.730 & 1.784 & 1.340 \\
    \textsc{Gemma-2-2B-IT} & 0.705 & 1.191 & 0.997 \\
    \textsc{Gemma-2-9B} & 0.709 & 2.216 & 1.685 \\
    \textsc{Gemma-2-9B-IT} & 0.638 & 1.234 & 0.978 \\
    \midrule
    \textsc{Pythia-410M} & 0.791 & 1.065 & 0.945 \\
    \bottomrule
    \end{tabular}
\end{table}

In this work, we collect three corpora from \texttt{Wiki}pedia, \texttt{Stack}Exchange, and \texttt{arXiv}. We only collect documents from 2025. 
More specifically,  we collect 3,366 documents for \texttt{Wiki}, 6,001 for \texttt{Stack}Exchange and  44,384 documents for \texttt{arXiv}.
These datasets are split into training, validation, and testing sets, in an 8:1:1 ratio.
For \texttt{arXiv}, we utilize only the paper abstracts, while for \texttt{Stack}Exchange, we use only the questions. Consequently, the average document length is 163 words for \texttt{arXiv}, 250 words for \texttt{Stack}Exchange, and 1,502 words for \texttt{Wiki}pedia.

We exclude \textsc{Gemma} models from our perplexity agreement experiments, as \texttt{lm-eval-harness} provides unreliable perplexity measurements for \textsc{Gemma} models\footnote{
See discussion at \url{https://github.com/huggingface/transformers/issues/29250}.
}.
We report the bits per byte (BPB) for the \textsc{Gemma} models in Table~\ref{app_tab:perplexity}.
While the BPB values for \textsc{Gemma} on \texttt{arXiv} (the dataset with the shortest average sequence length) are mostly reasonable, the performance on \texttt{Stack}Exchange and \texttt{Wiki}pedia is notably worse, even compared to smaller models like \textsc{Pythia-410M}.

This anomaly stems from how \texttt{lm-eval-harness} handles long sequences via a rolling window mechanism. 
Unlike other models, \textsc{Gemma} requires every input sequence to begin with the \texttt{BOS} token. If this constraint is not met, perplexity degrades significantly.
Consequently, when processing long sequences that are chunked into multiple windows, \textsc{Gemma}’s performance degrades.

% \clearpage
\subsection{PC1 Score under Train-before-Test}
\label{app:scale}
\begin{figure}[t]
    \centering
    \includegraphics[width=1.0\linewidth]{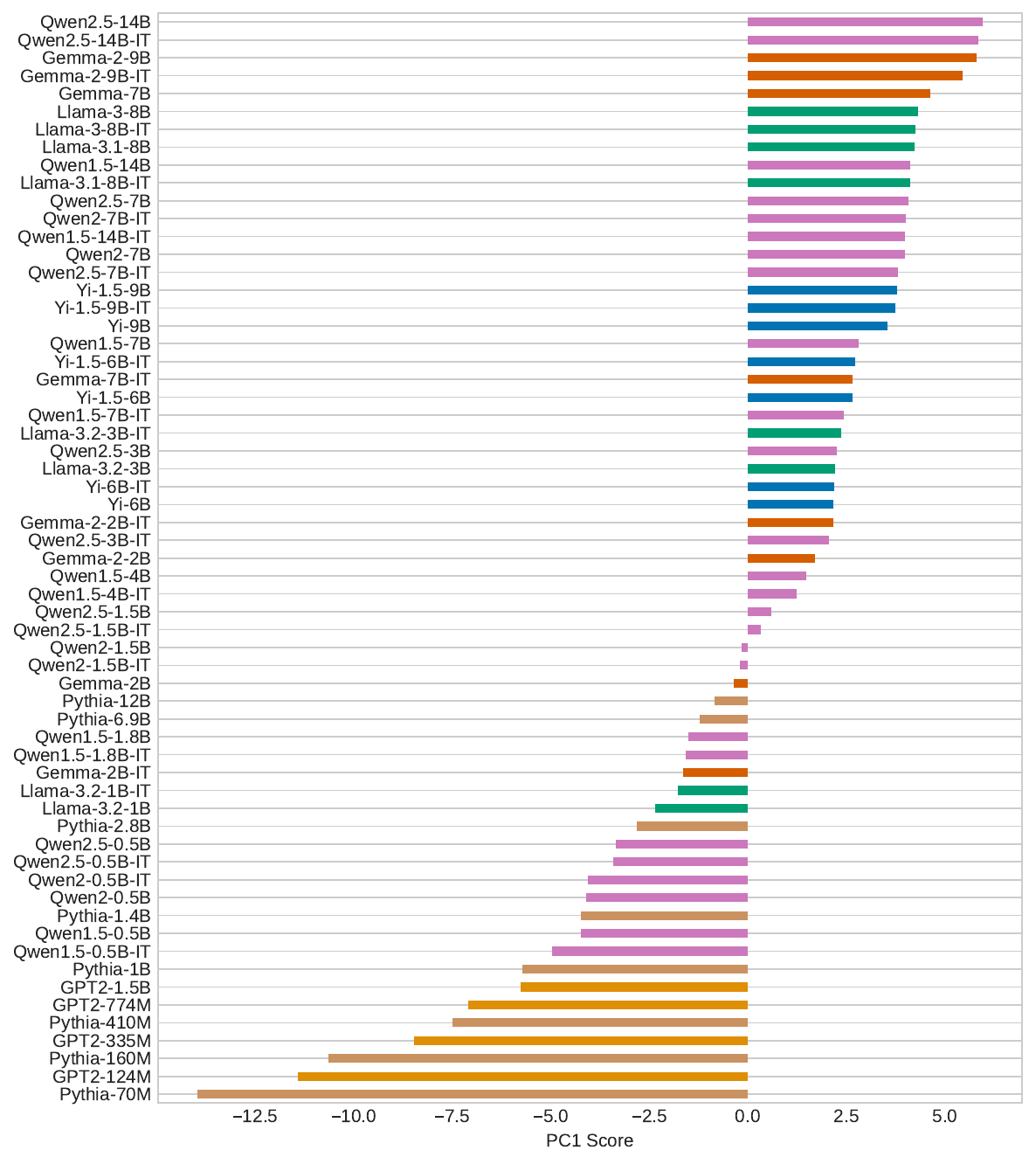}
    \caption{PC1 scores under train-before-test.}
    \label{app_fig:pc1_score}
\end{figure}

We plot the PC1 scores under train-before-test in Figure~\ref{app_fig:pc1_score}. 
We also provide the pre-training compute details for models with publicly available training token counts, as shown in Table~\ref{tab:flops}.

\begin{table}[t]
    \caption{
    The models used in Figure~\ref{fig:pc1_scale}.
    The number of training tokens of these models is publicly available.
    We compute the number of pre-training FLOPs as $6 \times \text{\#Parameters} \times \text{\#Tokens}$.
    }
    \label{tab:flops}
    \centering
    \begin{tabular}{cccc}
    \toprule
                   Model &  \#Parameters (B) &  \#Tokens (B) &  \#FLOPs (10\textasciicircum 18) \\
    \midrule
         Llama-3-8B &             8.03 &      15000.0 &       722700.00 \\
      Llama-3-8B-IT &             8.03 &      15000.0 &       722700.00 \\
            Llama-3.1-8B &             8.03 &      15000.0 &       722700.00 \\
         Llama-3.1-8B-IT &             8.03 &      15000.0 &       722700.00 \\
            Llama-3.2-3B &             3.21 &       9000.0 &       173340.00 \\
         Llama-3.2-3B-IT &             3.21 &       9000.0 &       173340.00 \\
            Qwen1.5-0.5B &             0.62 &       2400.0 &         8928.00 \\
            Qwen1.5-1.8B &             1.84 &       2400.0 &        26496.00 \\
              Qwen1.5-4B &             3.95 &       2400.0 &        56880.00 \\
              Qwen1.5-7B &             7.72 &       4000.0 &       185280.00 \\
             Qwen1.5-14B &            14.20 &       4000.0 &       340800.00 \\
         Qwen1.5-0.5B-IT &             0.62 &       2400.0 &         8928.00 \\
         Qwen1.5-1.8B-IT &             1.84 &       2400.0 &        26496.00 \\
           Qwen1.5-4B-IT &             3.95 &       2400.0 &        56880.00 \\
           Qwen1.5-7B-IT &             7.72 &       4000.0 &       185280.00 \\
          Qwen1.5-14B-IT &            14.20 &       4000.0 &       340800.00 \\
                Gemma-7B &             8.54 &       6000.0 &       307440.00 \\
             Gemma-7B-IT &             8.54 &       6000.0 &       307440.00 \\
              Gemma-2-2B &             2.61 &       2000.0 &        31320.00 \\
           Gemma-2-2B-IT &             2.61 &       2000.0 &        31320.00 \\
              Gemma-2-9B &             9.24 &       8000.0 &       443520.00 \\
           Gemma-2-9B-IT &             9.24 &       8000.0 &       443520.00 \\
              Pythia-70M &             0.07 &        300.0 &          126.00 \\
             Pythia-160M &             0.16 &        300.0 &          288.00 \\
             Pythia-410M &             0.41 &        300.0 &          738.00 \\
               Pythia-1B &             1.00 &        300.0 &         1800.00 \\
             Pythia-1.4B &             1.40 &        300.0 &         2520.00 \\
             Pythia-2.8B &             2.80 &        300.0 &         5040.00 \\
             Pythia-6.9B &             6.90 &        300.0 &        12420.00 \\
              Pythia-12B &            12.00 &        300.0 &        21600.00 \\
                   Yi-6B &             6.06 &       3000.0 &       109080.00 \\
                Yi-6B-IT &             6.06 &       3000.0 &       109080.00 \\
                   Yi-9B &             8.83 &       3800.0 &       201324.00 \\
               Yi-1.5-6B &             6.06 &       3600.0 &       130896.00 \\
            Yi-1.5-6B-IT &             6.06 &       3600.0 &       130896.00 \\
               Yi-1.5-9B &             8.83 &       3600.0 &       190728.00 \\
            Yi-1.5-9B-IT &             8.83 &       3600.0 &       190728.00 \\
    \bottomrule
    \end{tabular}
\end{table}

\clearpage
\section{Accounting for Statistical Significance}
\subsection{Ranking Alignment in Figure~\ref{fig:banner}}
\label{app:banner}
We plot the rankings of 61 language models on two question-answering benchmarks: Natural Questions Open and ARC Challenge in Figure~\ref{fig:banner}.
We greedily align each ranking as much as possible without violating confidence intervals, thus revealing only those ranking changes that are statistically significant.
See Algorithm~\ref{alg:banner} for more details.

\subsection{Downstream Ranking Agreement}
We additionally supplement the experiments presented in the main text by modifying the ranking correlation metric to account for statistical significance in the benchmark evaluations.
% \paragraph{Modifying Kendall's $\tau$ to account for statistical significance in benchmark evaluations.} 
Specifically, we use Kendall's $\tau$-b~\citep{kendall1945treatment}, which adjusts for ties in rankings. We consider two models tied on a given benchmark if their performance difference is not statistically significant at the 95\% confidence level. We assess statistical significance using a t-test based on the standard error of the mean performances.

\begin{figure}[t]
    \centering
    \includegraphics[width=0.99\linewidth]{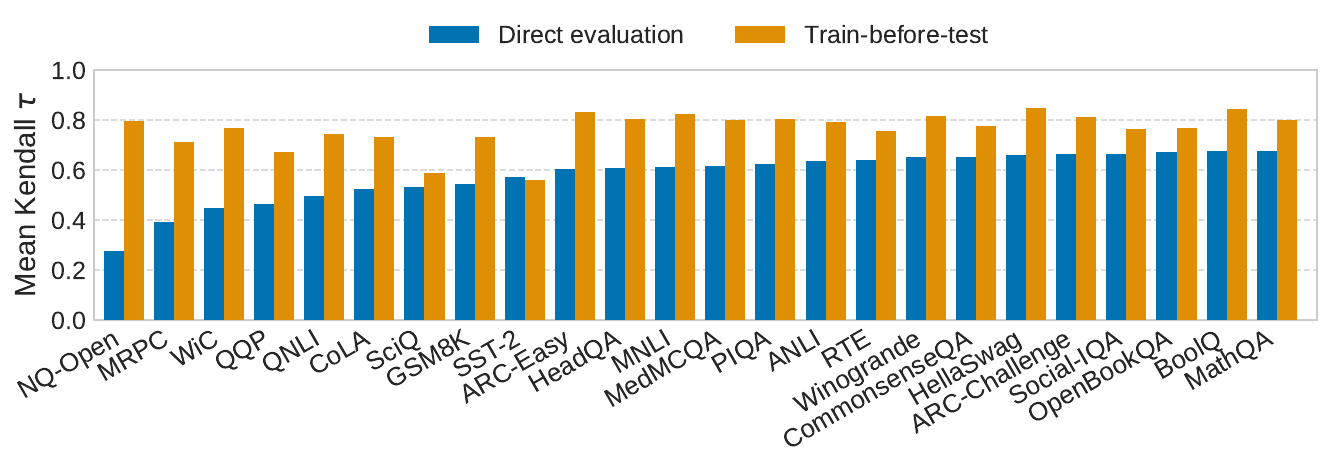}
    \caption{
    Mean ranking agreement between each benchmark and all others, measured by Kendall's $tau$-b, \emph{with non-statistically significant performance differences being treated as ties}. We calculate Kendall’s~$\tau$-b between each benchmark and every other one, and then average. Compared to direct evaluation, train-before-test consistently improves ranking agreement--often by a large margin. 
    On average, the overall average Kendall’s $\tau$ is 0.58 for direct evaluation and 0.77 for train-before-test.
    }
    \label{app_fig:downstream_consistency_ric}
\end{figure}

\begin{figure}
\centering
\begin{subfigure}{.49\textwidth}
  \centering
  \includegraphics[width=\linewidth]{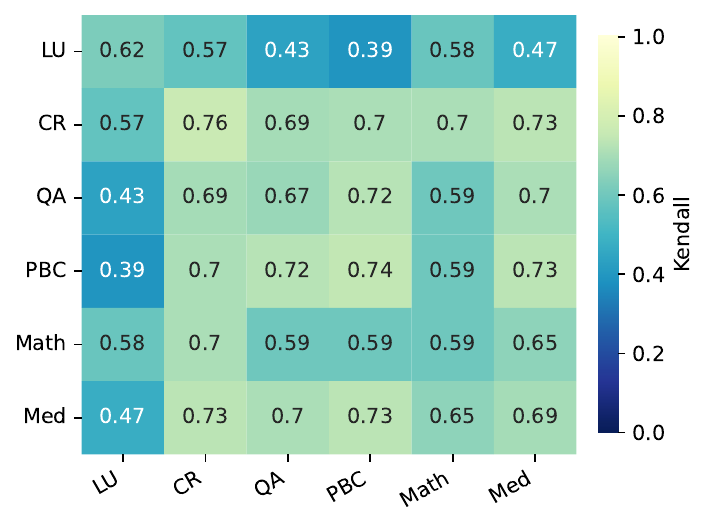}
  \caption{Direct evaluation.}
\end{subfigure}%
\begin{subfigure}{.49\textwidth}
  \centering
  \includegraphics[width=\linewidth]{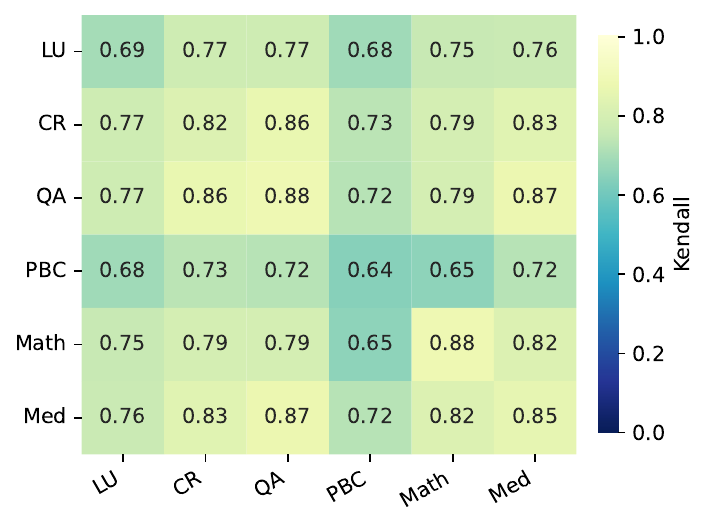}
  \caption{Train-before-test.}
\end{subfigure}
\caption{
Cross-category ranking agreement for direct evaluation (left) and train-before-test (right), measured by Kendall's $tau$-b, \emph{with non-statistically significant performance differences being treated as ties}.
We consider language understanding (LU), commonsense reasoning (CR), question answering (QA), physics/biology/chemistry (PBC), math (Math), and medicine (Med) categories. 
Kendall’s $\tau$-b is averaged across all pairs of benchmarks that belong to two given categories. 
The diagonal represents the intra-category agreement and the others represent the inter-category agreement.
train-before-test improves both intra- and inter-category ranking agreement in all instances.
}
\label{app_fig:cross_category_consistency_ric}
\end{figure}

\begin{figure}[t]
    \centering
    \includegraphics[width=0.99\linewidth]{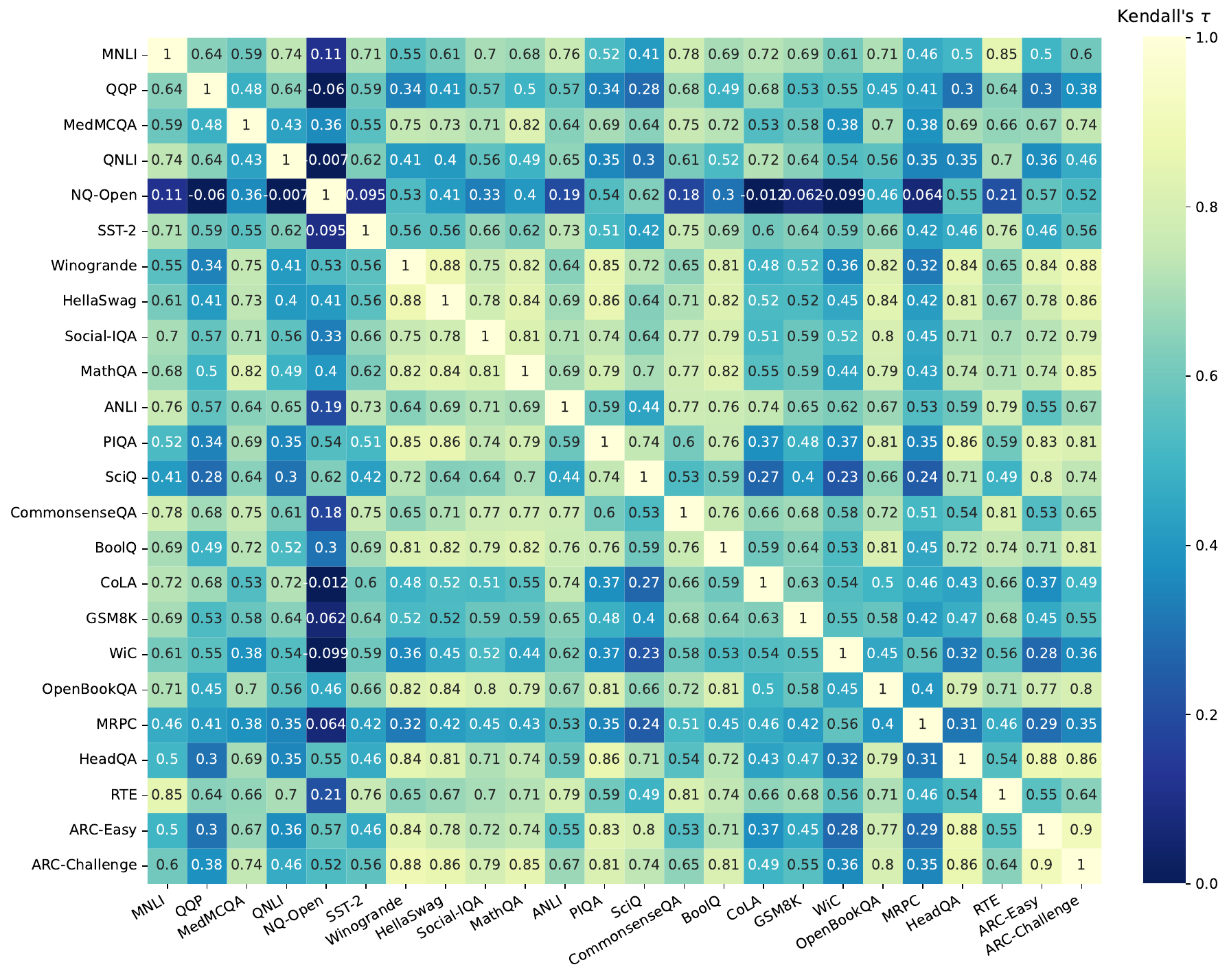}
    \caption{
    Cross benchmark ranking agreement under direct evaluation, measured by Kendall's $tau$-b with insignificant model comparisons treated as ties. 
    }
    \label{app_fig:detail_direct_eval_ric}
\end{figure}

\begin{figure}[t]
    \centering
    \includegraphics[width=0.99\linewidth]{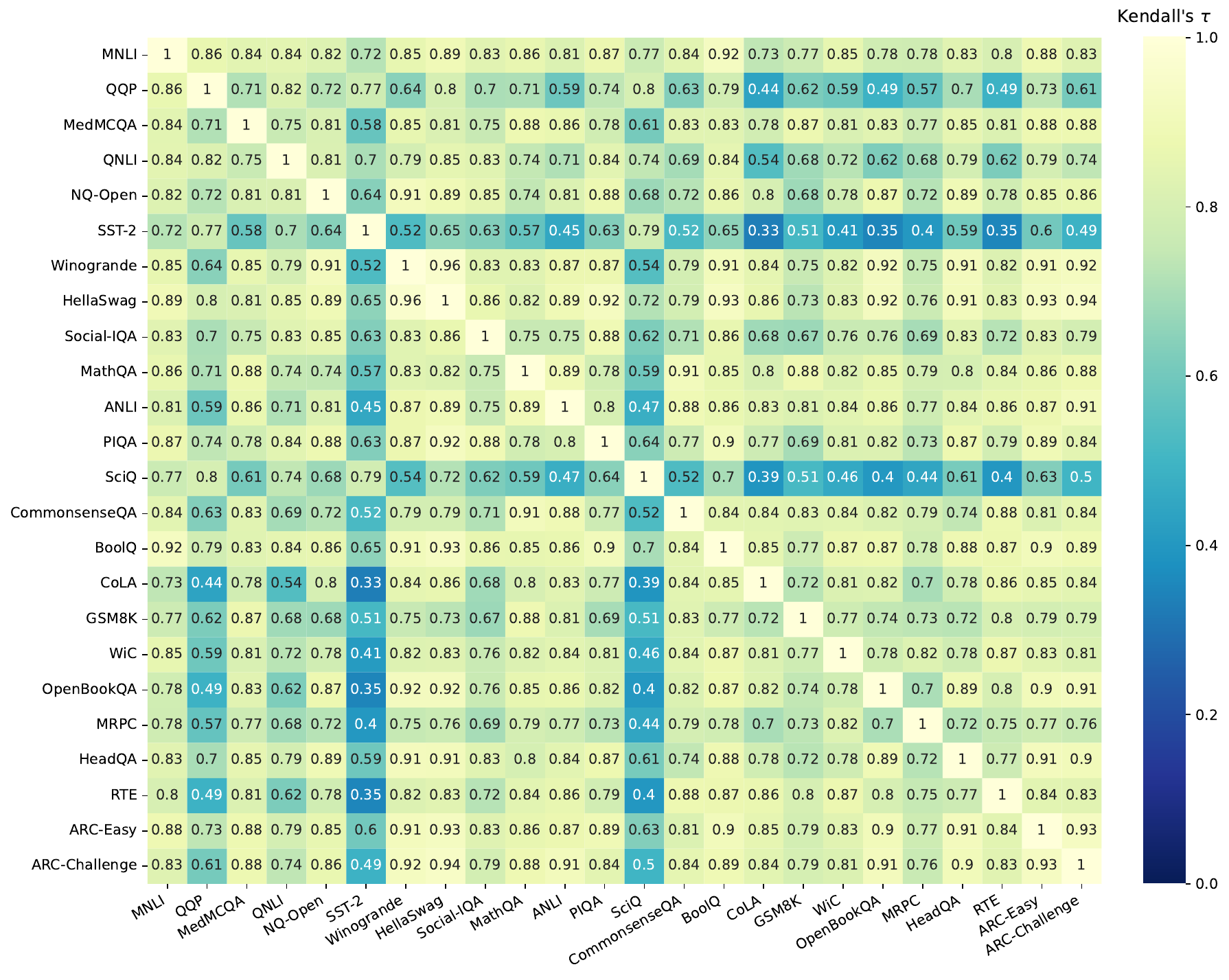}
    \caption{
    Cross benchmark ranking agreement under train-before-test, measured by Kendall's $tau$-b with insignificant model comparisons treated as ties. 
    }
    \label{app_fig:detail_train_before_test_ric}
\end{figure}

We reproduce the ranking correlation figures of the main text using the modified Kendall's $\tau$ which treats non-statistically significant performance differences as ties. See  Figure~\ref{app_fig:downstream_consistency_ric} and \ref{app_fig:cross_category_consistency_ric}; as well as Figure~\ref{app_fig:detail_direct_eval_ric} and Figure~\ref{app_fig:detail_train_before_test_ric} for more detailed results. We observe that accounting for statistical significance in models' performance differences leads to slightly higher ranking correlations, as measured by Kendall's $\tau$-b. For direct evaluation, average agreement increases from 0.52 to 0.58. For train-before-test, average agreement increases from 0.76 to 0.77. Therefore, train-before-test continues to lead to large improvements in raking agreement (from Kendall's $\tau$-b 0.58 to 0.77).

\clearpage
\begin{algorithm}[t]
\caption{build\_partial\_order(scores, stderrs)}
\KwIn{Model performance scores and standard errors}
\KwOut{Directed graph $G$ representing significant model orderings}

Initialize graph $G$ with models as nodes \\
\ForEach{pair of distinct models $(m_1, m_2)$}{
    \If{$m_1$ is significantly better than $m_2$}{
        Add directed edge $(m_1 \rightarrow m_2)$ to $G$
    }
}
\Return $G$
\end{algorithm}

\begin{algorithm}[t]
\caption{parallel\_greedy\_rank(models, $G_1$, $G_2$, score$_1$, score$_2$)}
\KwIn{List of models, two directed graphs $G_1$, $G_2$, and two score series}
\KwOut{Two lists representing the parallel ranking order for each task}

Initialize \texttt{vanillaRank$_1$}, $\leftarrow$ rankdata(score$_1$),
\texttt{vanillaRank$_2$} $\leftarrow$  rankdata(score$_2$) \\

Initialize \texttt{available$_1$} and \texttt{available$_2$} as models with zero in-degree in $G_1$ and $G_2$ \\
Initialize empty lists \texttt{order$_1$}, \texttt{order$_2$} \\

\For{$i = 1$ \KwTo number of models}{
    Initialize empty list \texttt{pairs} \\
    \ForEach{$m_1$ in \texttt{available$_1$}}{
        \ForEach{$m_2$ in \texttt{available$_2$}}{
            Compute cost for pair $(m_1, m_2)$ based on: \\
            \Indp
            (1) Placement of $m_1$ in \texttt{order$_2$} and $m_2$ in \texttt{order$_1$} \\
            (2) Whether $m_1 = m_2$ (prefer matching) \\
            (3) Combined vanilla ranks: \texttt{vanillaRank$_2$}[$m_1$] + \texttt{vanillaRank$_1$}[$m_2$] \\
            \Indm
            Append $(cost, m_1, m_2)$ to \texttt{pairs}
        }
    }
    Sort \texttt{pairs} by cost (ascending) \\
    Select $(m_1, m_2)$ with minimal cost \\
    Append $m_1$ to \texttt{order$_1$}, $m_2$ to \texttt{order$_2$} \\
    Remove $m_1$ from $G_1$ and update \texttt{available$_1$} \\
    Remove $m_2$ from $G_2$ and update \texttt{available$_2$} \\
}
\Return \texttt{order$_1$}, \texttt{order$_2$}
\end{algorithm}

\begin{algorithm}[t]
\caption{rank\_models(score$_1$, stderr$_1$, score$_2$, stderr$_2$)}
\label{alg:banner}
\KwIn{Scores and standard errors for two tasks}
\KwOut{Parallel rankings for both tasks}

\Indp
$G_1 \leftarrow$ \texttt{build\_partial\_order}(score$_1$, stderr$_1$) \\
$G_2 \leftarrow$ \texttt{build\_partial\_order}(score$_2$, stderr$_2$) \\
$(order_1, order_2) \leftarrow$ \texttt{parallel\_greedy\_rank}(\texttt{models}, $G_1$, $G_2$, score$_1$, score$_2$) \\
% Assign ranks to models based on their position in each order: \\
% \Indp
\texttt{rank$_1$}[$m$] $=$ position of $m$ in $order_1$ (starting from 1) \\
\texttt{rank$_2$}[$m$] $=$ position of $m$ in $order_2$ (starting from 1) \\
% \Indm
\Return \texttt{rank$_1$}, \texttt{rank$_2$}
\Indm
\end{algorithm}

\clearpage

%% file: main_arxiv.bbl
\begin{thebibliography}{10}

\bibitem{albadawy2018deep}
Ehab~A AlBadawy, Ashirbani Saha, and Maciej~A Mazurowski.
\newblock Deep learning for segmentation of brain tumors: Impact of cross-institutional training and testing.
\newblock {\em Medical physics}, 45(3):1150--1158, 2018.

\bibitem{albalaksurvey}
Alon Albalak, Yanai Elazar, Sang~Michael Xie, Shayne Longpre, Nathan Lambert, Xinyi Wang, Niklas Muennighoff, Bairu Hou, Liangming Pan, Haewon Jeong, et~al.
\newblock A survey on data selection for language models.
\newblock {\em Transactions on Machine Learning Research}, 2025.

\bibitem{mathqa}
Aida Amini, Saadia Gabriel, Peter Lin, Rik Koncel-Kedziorski, Yejin Choi, and Hannaneh Hajishirzi.
\newblock Mathqa: Towards interpretable math word problem solving with operation-based formalisms, 2019.

\bibitem{arrow2012social}
Kenneth~J. Arrow.
\newblock {\em Social Choice and Individual Values}.
\newblock Wiley, 1951.

\bibitem{barbu2019objectnet}
Andrei Barbu, David Mayo, Julian Alverio, William Luo, Christopher Wang, Dan Gutfreund, Josh Tenenbaum, and Boris Katz.
\newblock Objectnet: A large-scale bias-controlled dataset for pushing the limits of object recognition models.
\newblock {\em Advances in neural information processing systems}, 32, 2019.

\bibitem{open-llm-leaderboard}
Edward Beeching, Clémentine Fourrier, Nathan Habib, Sheon Han, Nathan Lambert, Nazneen Rajani, Omar Sanseviero, Lewis Tunstall, and Thomas Wolf.
\newblock Open llm leaderboard.
\newblock \url{https://huggingface.co/spaces/HuggingFaceH4/open_llm_leaderboard}, 2023.

\bibitem{rte3}
Luisa Bentivogli, Ido Dagan, Hoa~Trang Dang, Danilo Giampiccolo, and Bernardo Magnini.
\newblock The fifth {PASCAL} recognizing textual entailment challenge.
\newblock In {\em Text Analysis Conference (TAC)}, 2009.

\bibitem{pythia}
Stella Biderman, Hailey Schoelkopf, Quentin~G. Anthony, Herbie Bradley, Kyle O'Brien, Eric Hallahan, Mohammad~Aflah Khan, Shivanshu Purohit, USVSN~Sai Prashanth, Edward Raff, Aviya Skowron, Lintang Sutawika, and Oskar van~der Wal.
\newblock Pythia: A suite for analyzing large language models across training and scaling.
\newblock {\em ArXiv}, abs/2304.01373, 2023.

\bibitem{piqa}
Yonatan Bisk, Rowan Zellers, Ronan~Le Bras, Jianfeng Gao, and Yejin Choi.
\newblock Piqa: Reasoning about physical commonsense in natural language, 2019.

\bibitem{brandfonbrener2024loss}
David Brandfonbrener, Nikhil Anand, Nikhil Vyas, Eran Malach, and Sham Kakade.
\newblock Loss-to-loss prediction: Scaling laws for all datasets.
\newblock {\em arXiv preprint arXiv:2411.12925}, 2024.

\bibitem{Brown2020LanguageMA}
Tom~B. Brown, Benjamin Mann, Nick Ryder, Melanie Subbiah, Jared Kaplan, Prafulla Dhariwal, Arvind Neelakantan, Pranav Shyam, Girish Sastry, Amanda Askell, Sandhini Agarwal, Ariel Herbert-Voss, Gretchen Krueger, Tom Henighan, Rewon Child, Aditya Ramesh, Daniel~M. Ziegler, Jeff Wu, Clemens Winter, Christopher Hesse, Mark Chen, Eric Sigler, Ma~teusz Litwin, Scott Gray, Benjamin Chess, Jack Clark, Christopher Berner, Sam McCandlish, Alec Radford, Ilya Sutskever, and Dario Amodei.
\newblock Language models are few-shot learners.
\newblock {\em ArXiv}, abs/2005.14165, 2020.

\bibitem{Burnell2023RevealingTS}
Ryan Burnell, Hank Hao, Andrew R.~A. Conway, and Jos{\'e}~Hern{\'a}ndez Orallo.
\newblock Revealing the structure of language model capabilities.
\newblock {\em ArXiv}, abs/2306.10062, 2023.

\bibitem{candela2009dataset}
J~Qui{\~n}onero Candela, Masashi Sugiyama, Anton Schwaighofer, and Neil~D Lawrence.
\newblock Dataset shift in machine learning.
\newblock {\em The MIT Press}, 1:5, 2009.

\bibitem{boolq}
Christopher Clark, Kenton Lee, Ming-Wei Chang, Tom Kwiatkowski, Michael Collins, and Kristina Toutanova.
\newblock Boolq: Exploring the surprising difficulty of natural yes/no questions, 2019.

\bibitem{allenai:arc}
Peter Clark, Isaac Cowhey, Oren Etzioni, Tushar Khot, Ashish Sabharwal, Carissa Schoenick, and Oyvind Tafjord.
\newblock Think you have solved question answering? try arc, the ai2 reasoning challenge.
\newblock {\em arXiv:1803.05457v1}, 2018.

\bibitem{gsm8k}
Karl Cobbe, Vineet Kosaraju, Mo~Bavarian, Mark Chen, Heewoo Jun, Lukasz Kaiser, Matthias Plappert, Jerry Tworek, Jacob Hilton, Reiichiro Nakano, Christopher Hesse, and John Schulman.
\newblock Training verifiers to solve math word problems.
\newblock {\em ArXiv}, abs/2110.14168, 2021.

\bibitem{rte}
Ido Dagan, Oren Glickman, and Bernardo Magnini.
\newblock The {PASCAL} recognising textual entailment challenge.
\newblock In {\em Machine learning challenges. evaluating predictive uncertainty, visual object classification, and recognising tectual entailment}, pages 177--190. Springer, 2006.

\bibitem{deng2009imagenet}
Jia Deng, Wei Dong, Richard Socher, Li-Jia Li, Kai Li, and Li~Fei-Fei.
\newblock Imagenet: A large-scale hierarchical image database.
\newblock In {\em 2009 IEEE conference on computer vision and pattern recognition}, pages 248--255. Ieee, 2009.

\bibitem{mrpc}
William~B Dolan and Chris Brockett.
\newblock Automatically constructing a corpus of sentential paraphrases.
\newblock In {\em Proceedings of the International Workshop on Paraphrasing}, 2005.

\bibitem{dominguez2024training}
Ricardo Dominguez-Olmedo, Florian~E Dorner, and Moritz Hardt.
\newblock Training on the test task confounds evaluation and emergence.
\newblock {\em arXiv preprint arXiv:2407.07890}, 2024.

\bibitem{benchmarkloss}
Zhengxiao Du, Aohan Zeng, Yuxiao Dong, and Jie Tang.
\newblock Understanding emergent abilities of language models from the loss perspective.
\newblock In {\em The Thirty-eighth Annual Conference on Neural Information Processing Systems}, 2024.

\bibitem{open-llm-leaderboard-v2}
Clémentine Fourrier, Nathan Habib, Alina Lozovskaya, Konrad Szafer, and Thomas Wolf.
\newblock Open llm leaderboard v2.
\newblock \url{https://huggingface.co/spaces/open-llm-leaderboard/open_llm_leaderboard}, 2024.

\bibitem{Gadre2023DataCompIS}
Samir~Yitzhak Gadre, Gabriel Ilharco, Alex Fang, Jonathan Hayase, Georgios Smyrnis, Thao Nguyen, Ryan Marten, Mitchell Wortsman, Dhruba Ghosh, Jieyu Zhang, Eyal Orgad, Rahim Entezari, Giannis Daras, Sarah Pratt, Vivek Ramanujan, Yonatan Bitton, Kalyani Marathe, Stephen Mussmann, Richard Vencu, Mehdi Cherti, Ranjay Krishna, Pang~Wei Koh, Olga Saukh, Alexander~J. Ratner, Shuran Song, Hannaneh Hajishirzi, Ali Farhadi, Romain Beaumont, Sewoong Oh, Alexandros~G. Dimakis, Jenia Jitsev, Yair Carmon, Vaishaal Shankar, and Ludwig Schmidt.
\newblock Datacomp: In search of the next generation of multimodal datasets.
\newblock {\em ArXiv}, abs/2304.14108, 2023.

\bibitem{Gadre2024LanguageMS}
Samir~Yitzhak Gadre, Georgios Smyrnis, Vaishaal Shankar, Suchin Gururangan, Mitchell Wortsman, Rulin Shao, Jean-Pierre Mercat, Alex Fang, Jeffrey Li, Sedrick~Scott Keh, Rui Xin, Marianna Nezhurina, Igor Vasiljevic, Jenia Jitsev, Alexandros~G. Dimakis, Gabriel Ilharco, Shuran Song, Thomas Kollar, Yair Carmon, Achal Dave, Reinhard Heckel, Niklas Muennighoff, and Ludwig Schmidt.
\newblock Language models scale reliably with over-training and on downstream tasks.
\newblock {\em ArXiv}, abs/2403.08540, 2024.

\bibitem{ganguli2022predictability}
Deep Ganguli, Danny Hernandez, Liane Lovitt, Amanda Askell, Yuntao Bai, Anna Chen, Tom Conerly, Nova Dassarma, Dawn Drain, Nelson Elhage, et~al.
\newblock Predictability and surprise in large generative models.
\newblock In {\em Proceedings of the 2022 ACM Conference on Fairness, Accountability, and Transparency}, pages 1747--1764, 2022.

\bibitem{eval-harness}
Leo Gao, Jonathan Tow, Baber Abbasi, Stella Biderman, Sid Black, Anthony DiPofi, Charles Foster, Laurence Golding, Jeffrey Hsu, Alain Le~Noac'h, Haonan Li, Kyle McDonell, Niklas Muennighoff, Chris Ociepa, Jason Phang, Laria Reynolds, Hailey Schoelkopf, Aviya Skowron, Lintang Sutawika, Eric Tang, Anish Thite, Ben Wang, Kevin Wang, and Andy Zou.
\newblock A framework for few-shot language model evaluation, 12 2023.

\bibitem{Anil2023GeminiAF}
Gemini.
\newblock Gemini: A family of highly capable multimodal models.
\newblock {\em arXiv}, 2023.

\bibitem{gemma}
Gemma, Morgane Riviere, Shreya Pathak, Pier~Giuseppe Sessa, Cassidy Hardin, Surya Bhupatiraju, L{\'e}onard Hussenot, Thomas Mesnard, Bobak Shahriari, Alexandre Ram{\'e}, et~al.
\newblock Gemma 2: Improving open language models at a practical size.
\newblock {\em arXiv preprint arXiv:2408.00118}, 2024.

\bibitem{ghosh2024onebench}
Adhiraj Ghosh, Sebastian Dziadzio, Ameya Prabhu, Vishaal Udandarao, Samuel Albanie, and Matthias Bethge.
\newblock Onebench to test them all: Sample-level benchmarking over open-ended capabilities.
\newblock {\em arXiv preprint arXiv:2412.06745}, 2024.

\bibitem{rte2}
Danilo Giampiccolo, Bernardo Magnini, Ido Dagan, and Bill Dolan.
\newblock The third {PASCAL} recognizing textual entailment challenge.
\newblock In {\em Proceedings of the ACL-PASCAL workshop on textual entailment and paraphrasing}, pages 1--9. Association for Computational Linguistics, 2007.

\bibitem{llama}
Aaron Grattafiori, Abhimanyu Dubey, Abhinav Jauhri, Abhinav Pandey, Abhishek Kadian, Ahmad Al-Dahle, Aiesha Letman, Akhil Mathur, Alan Schelten, Alex Vaughan, et~al.
\newblock The llama 3 herd of models.
\newblock {\em arXiv preprint arXiv:2407.21783}, 2024.

\bibitem{Guha2025OpenThoughtsDR}
Etash~Kumar Guha, Ryan Marten, Sedrick~Scott Keh, Negin Raoof, Georgios Smyrnis, Hritik Bansal, Marianna Nezhurina, Jean-Pierre Mercat, Trung Vu, Zayne Sprague, Ashima Suvarna, Ben Feuer, Liangyu Chen, Zaid Khan, Eric Frankel, Sachin Grover, Caroline Choi, Niklas Muennighoff, Shiye Su, Wanjia Zhao, John Yang, Shreyas Pimpalgaonkar, Kartik Sharma, Charlie Cheng-Jie Ji, Yichuan Deng, Sarah Pratt, Vivek Ramanujan, Jon Saad-Falcon, Jeffrey Li, Achal Dave, Alon Albalak, Kushal Arora, Blake Wulfe, Chinmay Hegde, Greg Durrett, Sewoong Oh, Mohit Bansal, Saadia Gabriel, Aditya Grover, Kai-Wei Chang, Vaishaal Shankar, Aaron Gokaslan, Mike~A. Merrill, Tatsunori Hashimoto, Yejin Choi, Jenia Jitsev, Reinhard Heckel, Maheswaran Sathiamoorthy, Alexandros~G. Dimakis, and Ludwig Schmidt.
\newblock Openthoughts: Data recipes for reasoning models.
\newblock {\em ArXiv}, abs/2506.04178, 2025.

\bibitem{guo2025deepseek}
Daya Guo, Dejian Yang, Haowei Zhang, Junxiao Song, Ruoyu Zhang, Runxin Xu, Qihao Zhu, Shirong Ma, Peiyi Wang, Xiao Bi, et~al.
\newblock Deepseek-r1: Incentivizing reasoning capability in llms via reinforcement learning.
\newblock {\em arXiv preprint arXiv:2501.12948}, 2025.

\bibitem{hardt2025emerging}
Moritz Hardt.
\newblock The emerging science of machine learning benchmarks.
\newblock Online at \url{https://mlbenchmarks.org}, 2025.
\newblock Manuscript.

\bibitem{hardt2022patterns}
Moritz Hardt and Benjamin Recht.
\newblock {\em Patterns, predictions, and actions: Foundations of machine learning}.
\newblock Princeton University Press, 2022.

\bibitem{mmlu}
Dan Hendrycks, Collin Burns, Steven Basart, Andy Zou, Mantas Mazeika, Dawn~Xiaodong Song, and Jacob Steinhardt.
\newblock Measuring massive multitask language understanding.
\newblock {\em ArXiv}, abs/2009.03300, 2020.

\bibitem{Hoffmann2022TrainingCL}
Jordan Hoffmann, Sebastian Borgeaud, Arthur Mensch, Elena Buchatskaya, Trevor Cai, Eliza Rutherford, Diego de~Las~Casas, Lisa~Anne Hendricks, Johannes Welbl, Aidan Clark, Tom Hennigan, Eric Noland, Katie Millican, George van~den Driessche, Bogdan Damoc, Aurelia Guy, Simon Osindero, Karen Simonyan, Erich Elsen, Jack~W. Rae, Oriol Vinyals, and L.~Sifre.
\newblock Training compute-optimal large language models.
\newblock {\em ArXiv}, abs/2203.15556, 2022.

\bibitem{Hu2021LoRALA}
J.~Edward Hu, Yelong Shen, Phillip Wallis, Zeyuan Allen-Zhu, Yuanzhi Li, Shean Wang, and Weizhu Chen.
\newblock Lora: Low-rank adaptation of large language models.
\newblock {\em ArXiv}, abs/2106.09685, 2021.

\bibitem{huan2025does}
Maggie Huan, Yuetai Li, Tuney Zheng, Xiaoyu Xu, Seungone Kim, Minxin Du, Radha Poovendran, Graham Neubig, and Xiang Yue.
\newblock Does math reasoning improve general llm capabilities? understanding transferability of llm reasoning.
\newblock {\em arXiv preprint arXiv:2507.00432}, 2025.

\bibitem{kaplan2020scaling}
Jared Kaplan, Sam McCandlish, Tom Henighan, Tom~B Brown, Benjamin Chess, Rewon Child, Scott Gray, Alec Radford, Jeffrey Wu, and Dario Amodei.
\newblock Scaling laws for neural language models.
\newblock {\em arXiv preprint arXiv:2001.08361}, 2020.

\bibitem{kendall1938new}
Maurice~G Kendall.
\newblock A new measure of rank correlation.
\newblock {\em Biometrika}, 30(1-2):81--93, 1938.

\bibitem{kendall1945treatment}
Maurice~G Kendall.
\newblock The treatment of ties in ranking problems.
\newblock {\em Biometrika}, 33(3):239--251, 1945.

\bibitem{kornblith2019better}
Simon Kornblith, Jonathon Shlens, and Quoc~V Le.
\newblock Do better imagenet models transfer better?
\newblock In {\em Proceedings of the IEEE/CVF conference on computer vision and pattern recognition}, pages 2661--2671, 2019.

\bibitem{nq_open}
Tom Kwiatkowski, Jennimaria Palomaki, Olivia Redfield, Michael Collins, Ankur Parikh, Chris Alberti, Danielle Epstein, Illia Polosukhin, Jacob Devlin, Kenton Lee, Kristina Toutanova, Llion Jones, Matthew Kelcey, Ming-Wei Chang, Andrew~M. Dai, Jakob Uszkoreit, Quoc Le, and Slav Petrov.
\newblock Natural questions: A benchmark for question answering research.
\newblock {\em Transactions of the Association for Computational Linguistics}, 7:452--466, 2019.

\bibitem{Lester2021ThePO}
Brian Lester, Rami Al-Rfou, and Noah Constant.
\newblock The power of scale for parameter-efficient prompt tuning.
\newblock In {\em Conference on Empirical Methods in Natural Language Processing}, 2021.

\bibitem{winograd}
Hector~J Levesque, Ernest Davis, and Leora Morgenstern.
\newblock The {W}inograd schema challenge.
\newblock In {\em {AAAI} Spring Symposium: Logical Formalizations of Commonsense Reasoning}, volume~46, page~47, 2011.

\bibitem{liang2022holistic}
Percy Liang, Rishi Bommasani, Tony Lee, Dimitris Tsipras, Dilara Soylu, Michihiro Yasunaga, Yian Zhang, Deepak Narayanan, Yuhuai Wu, Ananya Kumar, et~al.
\newblock Holistic evaluation of language models.
\newblock {\em Annals of the New York Academy of Sciences}, 1525:140 -- 146, 2023.

\bibitem{liao2021we}
Thomas Liao, Rohan Taori, Inioluwa~Deborah Raji, and Ludwig Schmidt.
\newblock Are we learning yet? a meta review of evaluation failures across machine learning.
\newblock In {\em Thirty-fifth Conference on Neural Information Processing Systems Datasets and Benchmarks Track (Round 2)}, 2021.

\bibitem{liberman2010obituary}
Mark Liberman.
\newblock Obituary: Fred jelinek.
\newblock {\em Computational Linguistics}, 36(4):595--599, 2010.

\bibitem{Lin2024SelectingLL}
Haowei Lin, Baizhou Huang, Haotian Ye, Qinyu Chen, Zihao Wang, Sujian Li, Jianzhu Ma, Xiaojun Wan, James Zou, and Yitao Liang.
\newblock Selecting large language model to fine-tune via rectified scaling law.
\newblock {\em ArXiv}, abs/2402.02314, 2024.

\bibitem{liu2023same}
Hong Liu, Sang~Michael Xie, Zhiyuan Li, and Tengyu Ma.
\newblock Same pre-training loss, better downstream: Implicit bias matters for language models.
\newblock In {\em International Conference on Machine Learning}, pages 22188--22214. PMLR, 2023.

\bibitem{Lourie2025ScalingLA}
Nicholas Lourie, Michael~Y. Hu, and Kyunghyun Cho.
\newblock Scaling laws are unreliable for downstream tasks: A reality check.
\newblock {\em ArXiv}, abs/2507.00885, 2025.

\bibitem{lourie2025scaling}
Nicholas Lourie, Michael~Y Hu, and Kyunghyun Cho.
\newblock Scaling laws are unreliable for downstream tasks: A reality check.
\newblock {\em arXiv preprint arXiv:2507.00885}, 2025.

\bibitem{Magnusson2023PalomaAB}
Ian Magnusson, Akshita Bhagia, Valentin Hofmann, Luca Soldaini, A.~Jha, Oyvind Tafjord, Dustin Schwenk, Pete Walsh, Yanai Elazar, Kyle Lo, Dirk Groeneveld, Iz~Beltagy, Hanna Hajishirzi, Noah~A. Smith, Kyle Richardson, and Jesse Dodge.
\newblock Paloma: A benchmark for evaluating language model fit.
\newblock {\em ArXiv}, abs/2312.10523, 2023.

\bibitem{peft}
Sourab Mangrulkar, Sylvain Gugger, Lysandre Debut, Younes Belkada, Sayak Paul, and Benjamin Bossan.
\newblock Peft: State-of-the-art parameter-efficient fine-tuning methods.
\newblock \url{https://github.com/huggingface/peft}, 2022.

\bibitem{mayilvahananllms}
Prasanna Mayilvahanan, Thadd{\"a}us Wiedemer, Sayak Mallick, Matthias Bethge, and Wieland Brendel.
\newblock Llms on the line: Data determines loss-to-loss scaling laws.
\newblock In {\em Forty-second International Conference on Machine Learning}, 2025.

\bibitem{openbookqa}
Todor Mihaylov, Peter Clark, Tushar Khot, and Ashish Sabharwal.
\newblock Can a suit of armor conduct electricity? a new dataset for open book question answering.
\newblock In Ellen Riloff, David Chiang, Julia Hockenmaier, and Jun{'}ichi Tsujii, editors, {\em Proceedings of the 2018 Conference on Empirical Methods in Natural Language Processing}, pages 2381--2391, Brussels, Belgium, October-November 2018. Association for Computational Linguistics.

\bibitem{miller2020effect}
John Miller, Karl Krauth, Benjamin Recht, and Ludwig Schmidt.
\newblock The effect of natural distribution shift on question answering models.
\newblock In {\em International conference on machine learning}, pages 6905--6916. PMLR, 2020.

\bibitem{anli}
Yixin Nie, Adina Williams, Emily Dinan, Mohit Bansal, Jason Weston, and Douwe Kiela.
\newblock Adversarial nli: A new benchmark for natural language understanding, 2020.

\bibitem{openai2023GPT4}
OpenAI.
\newblock Gpt-4 technical report.
\newblock {\em arXiv}, 2023.

\bibitem{Owen2024HowPI}
David Owen.
\newblock How predictable is language model benchmark performance?
\newblock {\em ArXiv}, abs/2401.04757, 2024.

\bibitem{medmcqa}
Ankit Pal, Logesh~Kumar Umapathi, and Malaikannan Sankarasubbu.
\newblock Medmcqa : A large-scale multi-subject multi-choice dataset for medical domain question answering, 2022.

\bibitem{wic}
Mohammad~Taher Pilehvar and Jose Camacho-Collados.
\newblock Wic: the word-in-context dataset for evaluating context-sensitive meaning representations.
\newblock {\em arXiv preprint arXiv:1808.09121}, 2018.

\bibitem{gpt2}
Alec Radford, Jeff Wu, Rewon Child, David Luan, Dario Amodei, and Ilya Sutskever.
\newblock Language models are unsupervised multitask learners.
\newblock In {\em OpenAI Technical Report}, 2019.
\newblock OpenAI technical report.

\bibitem{qnli}
Pranav Rajpurkar, Jian Zhang, Konstantin Lopyrev, and Percy Liang.
\newblock {SQ}u{AD}: 100,000+ questions for machine comprehension of text.
\newblock In {\em Proceedings of EMNLP}, pages 2383--2392. Association for Computational Linguistics, 2016.

\bibitem{Ramesh2021ZeroShotTG}
Aditya Ramesh, Mikhail Pavlov, Gabriel Goh, Scott Gray, Chelsea Voss, Alec Radford, Mark Chen, and Ilya Sutskever.
\newblock Zero-shot text-to-image generation.
\newblock {\em ArXiv}, abs/2102.12092, 2021.

\bibitem{recht2019imagenetclassifiersgeneralizeimagenet}
Benjamin Recht, Rebecca Roelofs, Ludwig Schmidt, and Vaishaal Shankar.
\newblock Do imagenet classifiers generalize to imagenet?, 2019.

\bibitem{Ruan2024ObservationalSL}
Yangjun Ruan, Chris~J. Maddison, and Tatsunori~B. Hashimoto.
\newblock Observational scaling laws and the predictability of language model performance.
\newblock {\em ArXiv}, abs/2405.10938, 2024.

\bibitem{salaudeen2024imagenot}
Olawale Salaudeen and Moritz Hardt.
\newblock Imagenot: A contrast with imagenet preserves model rankings.
\newblock {\em arXiv preprint arXiv:2404.02112}, 2024.

\bibitem{Salaudeen2025MeasurementTM}
Olawale Salaudeen, Anka Reuel, Ahmed~M. Ahmed, Suhana Bedi, Zachary Robertson, Sudharsan Sundar, Ben Domingue, Angelina Wang, and Oluwasanmi Koyejo.
\newblock Measurement to meaning: A validity-centered framework for ai evaluation.
\newblock {\em ArXiv}, abs/2505.10573, 2025.

\bibitem{socialiqa}
Maarten Sap, Hannah Rashkin, Derek Chen, Ronan LeBras, and Yejin Choi.
\newblock Socialiqa: Commonsense reasoning about social interactions, 2019.

\bibitem{Shnitzer2023LargeLM}
Tal Shnitzer, Anthony Ou, M'irian Silva, Kate Soule, Yuekai Sun, Justin Solomon, Neil Thompson, and Mikhail Yurochkin.
\newblock Large language model routing with benchmark datasets.
\newblock {\em ArXiv}, abs/2309.15789, 2023.

\bibitem{sst2}
Richard Socher, Alex Perelygin, Jean Wu, Jason Chuang, Christopher~D Manning, Andrew Ng, and Christopher Potts.
\newblock Recursive deep models for semantic compositionality over a sentiment treebank.
\newblock In {\em Proceedings of EMNLP}, pages 1631--1642, 2013.

\bibitem{srivastava2022beyond}
Aarohi Srivastava, Abhinav Rastogi, Abhishek Rao, Abu Awal~Md Shoeb, Abubakar Abid, Adam Fisch, Adam~R Brown, Adam Santoro, Aditya Gupta, Adri{\`a} Garriga-Alonso, et~al.
\newblock Beyond the imitation game: Quantifying and extrapolating the capabilities of language models.
\newblock {\em ArXiv}, abs/2206.04615, 2022.

\bibitem{bbh}
Mirac Suzgun, Nathan Scales, Nathanael Scharli, Sebastian Gehrmann, Yi~Tay, Hyung~Won Chung, Aakanksha Chowdhery, Quoc~V. Le, Ed~H. Chi, Denny Zhou, and Jason Wei.
\newblock Challenging big-bench tasks and whether chain-of-thought can solve them.
\newblock In {\em Annual Meeting of the Association for Computational Linguistics}, 2022.

\bibitem{commonsenseqa}
Alon Talmor, Jonathan Herzig, Nicholas Lourie, and Jonathan Berant.
\newblock Commonsenseqa: A question answering challenge targeting commonsense knowledge, 2019.

\bibitem{taori2020measuring}
Rohan Taori, Achal Dave, Vaishaal Shankar, Nicholas Carlini, Benjamin Recht, and Ludwig Schmidt.
\newblock Measuring robustness to natural distribution shifts in image classification.
\newblock {\em Advances in Neural Information Processing Systems}, 33:18583--18599, 2020.

\bibitem{torralba2011unbiased}
Antonio Torralba and Alexei~A Efros.
\newblock Unbiased look at dataset bias.
\newblock In {\em CVPR 2011}, pages 1521--1528. IEEE, 2011.

\bibitem{Tsipras2020FromIT}
Dimitris Tsipras, Shibani Santurkar, Logan Engstrom, Andrew Ilyas, and Aleksander Madry.
\newblock From imagenet to image classification: Contextualizing progress on benchmarks.
\newblock In {\em International Conference on Machine Learning}, 2020.

\bibitem{headqa}
David Vilares and Carlos G{\'o}mez-Rodr{\'i}guez.
\newblock {HEAD}-{QA}: A healthcare dataset for complex reasoning.
\newblock In {\em Proceedings of the 57th Annual Meeting of the Association for Computational Linguistics}, pages 960--966, Florence, Italy, July 2019. Association for Computational Linguistics.

\bibitem{cola}
Alex Warstadt, Amanpreet Singh, and Samuel~R. Bowman.
\newblock Neural network acceptability judgments.
\newblock {\em arXiv preprint 1805.12471}, 2018.

\bibitem{weiemergent}
Jason Wei, Yi~Tay, Rishi Bommasani, Colin Raffel, Barret Zoph, Sebastian Borgeaud, Dani Yogatama, Maarten Bosma, Denny Zhou, Donald Metzler, et~al.
\newblock Emergent abilities of large language models.
\newblock {\em Transactions on Machine Learning Research}, 2022.

\bibitem{Weidinger2025TowardAE}
Laura Weidinger, Deborah Raji, Hanna Wallach, Margaret Mitchell, Angelina Wang, Olawale Salaudeen, Rishi Bommasani, Sayash Kapoor, Deep Ganguli, Sanmi Koyejo, and William Isaac.
\newblock Toward an evaluation science for generative ai systems.
\newblock {\em ArXiv}, abs/2503.05336, 2025.

\bibitem{sciq}
Johannes Welbl, Nelson~F. Liu, and Matt Gardner.
\newblock Crowdsourcing multiple choice science questions, 2017.

\bibitem{mnli}
Adina Williams, Nikita Nangia, and Samuel~R. Bowman.
\newblock A broad-coverage challenge corpus for sentence understanding through inference.
\newblock In {\em Proceedings of NAACL-HLT}, 2018.

\bibitem{xia2023training}
Mengzhou Xia, Mikel Artetxe, Chunting Zhou, Xi~Victoria Lin, Ramakanth Pasunuru, Danqi Chen, Luke Zettlemoyer, and Veselin Stoyanov.
\newblock Training trajectories of language models across scales.
\newblock In {\em The 61st Annual Meeting Of The Association For Computational Linguistics}, 2023.

\bibitem{Yadav2019ColdCT}
Chhavi Yadav and L{\'e}on Bottou.
\newblock Cold case: The lost mnist digits.
\newblock In {\em Neural Information Processing Systems}, 2019.

\bibitem{qwen}
An~Yang, Baosong Yang, Beichen Zhang, Binyuan Hui, Bo~Zheng, Bowen Yu, Chengyuan Li, Dayiheng Liu, Fei Huang, Haoran Wei, Huan Lin, Jian Yang, Jianhong Tu, Jianwei Zhang, Jianxin Yang, Jiaxi Yang, Jingren Zhou, Junyang Lin, Kai Dang, Keming Lu, Keqin Bao, Kexin Yang, Le~Yu, Mei Li, Mingfeng Xue, Pei Zhang, Qin Zhu, Rui Men, Runji Lin, Tianhao Li, Tianyi Tang, Tingyu Xia, Xingzhang Ren, Xuancheng Ren, Yang Fan, Yang Su, Yichang Zhang, Yu~Wan, Yuqiong Liu, Zeyu Cui, Zhenru Zhang, and Zihan Qiu.
\newblock Qwen2.5 technical report, 2025.

\bibitem{yi}
Alex Young, Bei Chen, Chao Li, Chengen Huang, Ge~Zhang, Guanwei Zhang, Heng Li, Jiangcheng Zhu, Jianqun Chen, Jing Chang, Kaidong Yu, Peng Liu, Qiang Liu, Shawn Yue, Senbin Yang, Shiming Yang, Tao Yu, Wen Xie, Wenhao Huang, Xiaohui Hu, Xiaoyi Ren, Xinyao Niu, Pengcheng Nie, Yuchi Xu, Yudong Liu, Yue Wang, Yuxuan Cai, Zhenyu Gu, Zhiyuan Liu, and Zonghong Dai.
\newblock Yi: Open foundation models by 01.ai.
\newblock {\em ArXiv}, abs/2403.04652, 2024.

\bibitem{hellaswag}
Rowan Zellers, Ari Holtzman, Yonatan Bisk, Ali Farhadi, and Yejin Choi.
\newblock Hellaswag: Can a machine really finish your sentence?, 2019.

\bibitem{Zeng2025LENSLLMUF}
Xinyue Zeng, Haohui Wang, Junhong Lin, Jun Wu, Tyler Cody, and Dawei Zhou.
\newblock Lensllm: Unveiling fine-tuning dynamics for llm selection.
\newblock {\em ArXiv}, abs/2505.03793, 2025.

\bibitem{Zhang2024WhenSM}
Biao Zhang, Zhongtao Liu, Colin Cherry, and Orhan Firat.
\newblock When scaling meets llm finetuning: The effect of data, model and finetuning method.
\newblock {\em ArXiv}, abs/2402.17193, 2024.

\bibitem{zhang2024inherent}
Guanhua Zhang and Moritz Hardt.
\newblock Inherent trade-offs between diversity and stability in multi-task benchmarks.
\newblock {\em arXiv preprint arXiv:2405.01719}, 2024.

\bibitem{Zhang2024TaskMA}
Jieyu Zhang, Weikai Huang, Zixian Ma, Oscar Michel, Dong He, Tanmay Gupta, Wei-Chiu Ma, Ali Farhadi, Aniruddha Kembhavi, and Ranjay Krishna.
\newblock Task me anything.
\newblock {\em ArXiv}, abs/2406.11775, 2024.

\end{thebibliography}
